\documentclass{article}

\pdfoutput=1

\usepackage[numbers,sort&compress]{natbib}
\usepackage[utf8]{inputenc} 
\usepackage[T1]{fontenc}    
\usepackage{hyperref}       
\usepackage{url}            
\usepackage{booktabs}       
\usepackage{amsfonts}       
\usepackage{nicefrac}       
\usepackage{microtype}      
\usepackage{xcolor}         
\usepackage{times}
\usepackage{soul}
\usepackage[small]{caption}
\usepackage{graphicx}
\usepackage{amsmath}
\usepackage{enumitem}
\usepackage{subcaption}
\usepackage{amsthm}
\usepackage{algorithm}
\usepackage{algorithmic}
\usepackage{multirow}
\usepackage{array}
\usepackage{colortbl}
\usepackage{tabularx}
\usepackage{bbm}
\usepackage[flushleft]{threeparttable}
\usepackage{makecell}
\usepackage{svg}
\usepackage{wrapfig}

\usepackage{pifont}
\usepackage{blindtext}
\usepackage{rotating}
\usepackage{adjustbox}

\usepackage[final]{neurips_2025}

\title{FuncGenFoil: Airfoil Generation and Editing Model \\ in Function Space}

\author{%
Jinouwen Zhang$^{1*}$\quad
Junjie Ren$^{1,2}$\quad
Qianhong Ma$^{1,8}$\quad
Jianyu Wu$^{1}$\quad
Aobo Yang$^{3}$\\
\textbf{Yan Lu}$^{1,4}$\quad
\textbf{Lu Chen}$^{1,5}$\quad
\textbf{Hairun Xie}$^{6,7}$\quad
\textbf{Jing Wang}$^{6,8}$\\
\textbf{Miao Zhang}$^6$\quad
\textbf{Wanli Ouyang}$^{1,4}$\quad
\textbf{Shixiang Tang}$^{1,4\star}$ \\
$^1$ Shanghai Artificial Intelligence Laboratory \quad $^2$ Fudan University \\
$^3$ Hong Kong University of Science and Technology \quad $^4$ The Chinese University of Hong Kong\\
$^5$ State Key Lab of CAD\&CG, Zhejiang University \\ 
$^6$ Shanghai Aircraft Design and Research Institute\\
$^7$ Innovation Academy for Microsatellites of CAS \quad $^8$ Shanghai Jiao Tong University\\
  $^{*}$\texttt{zhangjinouwen@pjlab.org.cn} \quad  $^{\star}$\texttt{tangshixiang@pjlab.org.cn} \\
}

\begin{document}

\maketitle

\begin{abstract}
Aircraft manufacturing is the jewel in the crown of industry, in which generating high-fidelity airfoil geometries with controllable and editable representations remains a fundamental challenge. Existing deep learning methods, which typically rely on predefined parametric representations (e.g., Bézier curves) or discrete point sets, face an inherent trade-off between expressive power and resolution adaptability.
To tackle this challenge, we introduce \textbf{FuncGenFoil}, a novel function-space generative model that directly reconstructs airfoil geometries as function curves. Our method inherits the advantages of arbitrary-resolution sampling and smoothness from parametric functions, as well as the strong expressiveness of discrete point-based representations.
Empirical evaluations demonstrate that \textbf{FuncGenFoil} improves upon state-of-the-art methods in airfoil generation, achieving a relative \textbf{74.4\%} reduction in label error and a \textbf{23.2\%} increase in diversity on the AF-200K dataset. Our results highlight the advantages of function-space modeling for aerodynamic shape optimization, offering a powerful and flexible framework for high-fidelity airfoil design. 
\end{abstract}

\section{Introduction}

The airfoil inverse design problem serves as a central aspect of aircraft manufacturing. Traditionally, given geometric requirements, engineers first select the most similar airfoils from well-known airfoil datasets (e.g., NACA~\cite{Cummings2015AppliedCA}) and rely on a trial-and-error strategy~\cite{sharma2021recent}. Considering the mission of the aircraft, an initial airfoil design that meets these conditions is preliminarily created. Then, through iterative rounds of physical analyses, such as aerodynamic and mechanical evaluations, the airfoil is optimized to achieve improved performance until the specified requirements are met. In practice, such direct design procedures are highly inefficient and time-consuming, often taking months. To minimize development and design time, as well as associated costs, automated design methods have been introduced as efficient alternatives in aircraft manufacturing engineering. In particular, machine learning-based design and optimization techniques have gained significant attention. However, before applying these algorithms to airfoil design, it is crucial to determine appropriate methods for representing airfoils within these algorithms.

\begin{figure}
    \centering
      \includegraphics[width=\linewidth]{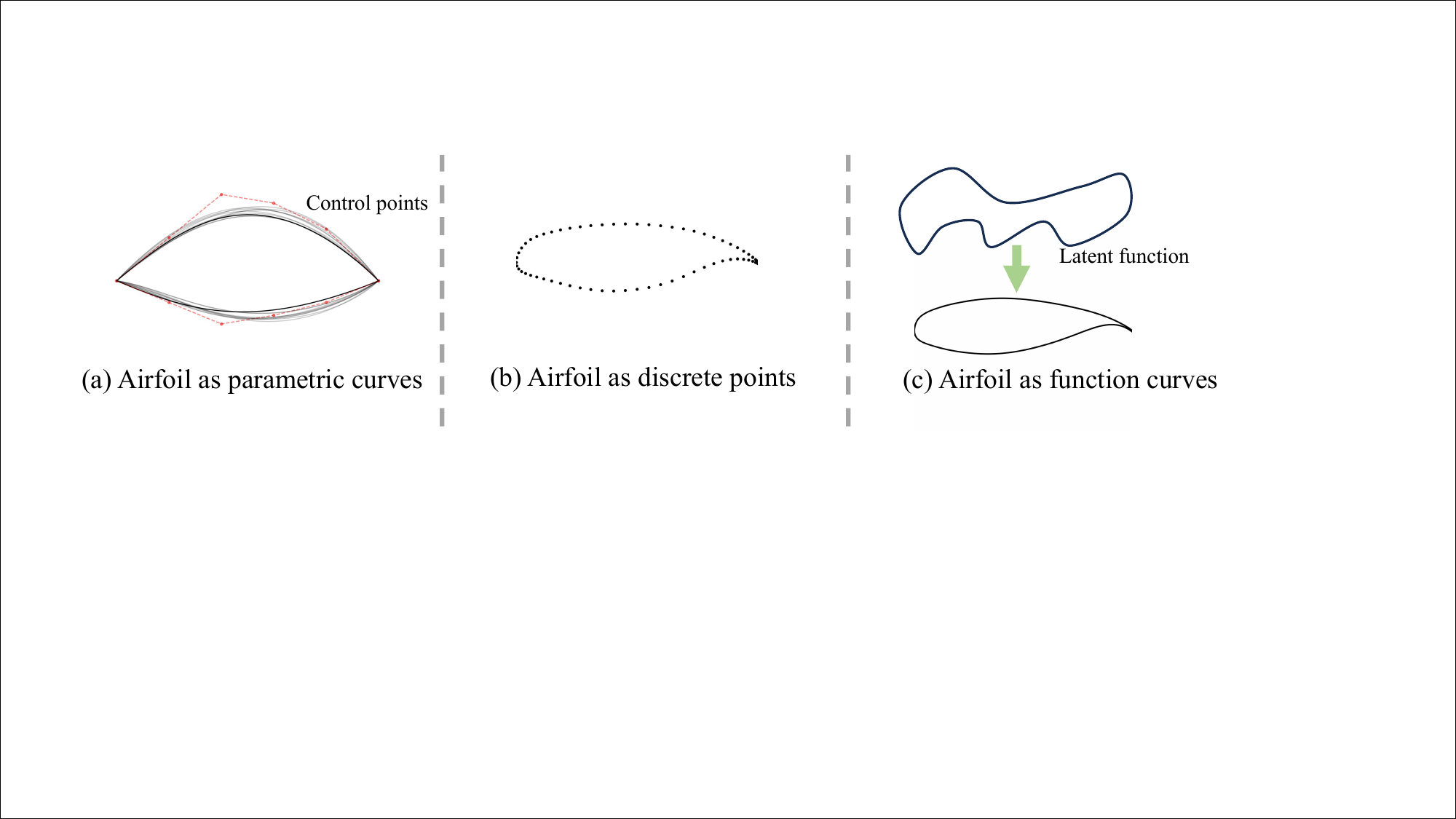}
      \caption{
      The conceptual difference between FuncGenFoil and previous airfoil representation methods. In many previous approaches, airfoils are represented either as parametric models, as shown in (a), or as discrete point-based models, as shown in (b). In contrast, (c) illustrates FuncGenFoil's approach, where an airfoil is treated as a continuous function mapped from a latent function, enabling a generative model in function space.}
      \label{fig:intro}
\end{figure}

Existing methods for airfoil representation can generally be divided into two categories: parametric-model-based approaches~\cite{xie2024parametric} and discrete-point-based methods~\cite{liu2024afbench,sekar2019inverse}. First, parametric-model-based methods predefine function families, \emph{e.g.,} Bézier curves~\cite{chen2021bezierganautomaticgenerationsmooth}, Hicks-Henne curves, and NURBS, and leverage mathematical optimization techniques or generative models to determine the parameters of these functions for generating a new airfoil. These methods rigorously preserve key geometric properties of the defined function families, \emph{e.g.,} higher-order smoothness. Furthermore, such functional representations of airfoils allow arbitrary sampling of control points in real manufacturing, given engineering precision constraints. Despite these benefits, parametric-model-based methods suffer from a significantly reduced design space; \emph{i.e.,} selecting a specific function family excludes other possible shapes, thus limiting the upper bound of airfoil design algorithms.
Second, discrete-point-based methods directly generate multiple points to represent airfoil shapes. These methods maximize the airfoil design space but cannot inherently maintain certain important mathematical properties, \emph{e.g.,} continuity. Furthermore, they cannot directly generate control points at arbitrary resolutions, as the number of generated points is typically fixed for each model after training.

To address the trade-offs between these two mainstream approaches, we ask: \emph{Can we design an algorithm that leverages the advantages of both?}

In this paper, we address this question by proposing \textbf{FuncGenFoil}, a novel function-space generative model for airfoil representation (see Figure~\ref{fig:intro}). Unlike previous data-driven generative methods in finite-dimensional spaces, \emph{e.g.}, cVAE~\cite{kingma2013auto}, cGAN~\cite{mirza2014conditional}, diffusion models~\cite{Ho2020}, or flow matching models~\cite{lipman2022flow}, which directly generate discrete point outputs, we leverage recent advances in diffusion and flow models defined in function space~\cite{pmlr-v206-lim23a,pmlr-v206-kerrigan23a,lim2025scorebaseddiffusionmodelsfunction,pidstrigach2023infinitedimensionaldiffusionmodels,franzese2024continuous,shi2024universal} to produce infinite-dimensional outputs as general continuous functions.
Simultaneously, our approach models airfoil geometry using Neural Operator architectures~\cite{anandkumar2019neural,li2021fourier,Kovachki2023,azizzadenesheli2024neural}, enabling the generation of diverse airfoils beyond the design spaces of predefined geometric function families, thanks to its general operator approximation capability.
Our method combines the advantages of both parametric-model-based and discrete-point-based approaches. Due to its intrinsic representation in function space, the generated airfoils are continuous, and both training and sampling can be performed at arbitrary resolutions, facilitating downstream optimization and manufacturing processes.

Specifically, we use the flow matching framework~\cite{lipman2022flow}, an improved alternative to diffusion models, and FNO~\cite{li2021fourier}, a resolution-free neural operator, as the backbone of our generative model to design FuncGenFoil. During training, we perturb the airfoil into a noise function through straight flows and let neural operators learn the airfoil's deconstructed velocity via flow matching. During inference, we reconstruct the airfoil by reversing the flow direction starting from a Gaussian process. Beyond generation, our method also supports airfoil editing, generating new airfoils from an original airfoil by conforming to certain geometric constraints. This can be accomplished by a few steps of fine-tuning with no additional data other than the original airfoil. Constraints can be provided in various forms, such as specifying contour points that the airfoil must pass through or setting geometric constraint parameters, such as thickness.

In summary, our main contributions are threefold: (1) We propose generating airfoil shapes in function space to achieve important properties for aircraft engineering, \emph{i.e.,} arbitrary-resolution control point sampling and maximal design space; (2) We design FuncGenFoil, the first controllable airfoil generative model in function space, which effectively incorporates neural operator architectures into the generative modeling framework; and (3) We further enhance FuncGenFoil with airfoil editing capabilities through minimal adaptations.
Experimental results indicate that our proposed method achieves state-of-the-art airfoil generation quality, reducing label error by 74.4\% and increasing diversity by 23.2\% on the AF-200K dataset, as validated through aerodynamic simulation analysis.
In addition, our method is the first to successfully perform airfoil editing by fixing or dragging points at arbitrary positions, achieving nearly zero MSE (less than $10^{-7}$).
\section{Related Works}

\noindent \textbf{Generative Models.}
Generative models based on score matching~\cite{ho2020denoising,Song2021a} and flow matching~\cite{lipman2023flow,tong2024improving,lipman2024flowmatchingguidecode} have significantly advanced machine learning, achieving state-of-the-art results in areas such as image generation~\cite{rombach2022high}, text generation~\cite{gat2024discrete}, and video generation~\cite{Ho2022video,polyak2024moviegencastmedia}. 
However, most of these models operate in finite-dimensional spaces and rely on fixed discretizations of the data. 
Such formulations hinder transferability across discretizations and neglect function-level constraints (e.g., continuity, smoothness), motivating the need for generative modeling in function space.

\noindent \textbf{Neural Architectures for Function Modeling.}
Designing neural architectures to handle function spaces remains a major research challenge. 
Standard networks typically assume fixed-size inputs, making them unsuitable for arbitrary resolutions. 
Implicit neural representations, such as SIREN~\cite{sitzmann2020implicit}, harness random Fourier features~\cite{rahimi2007random} to represent continuous and differentiable objects through position embeddings. 
Similarly, NeRF~\cite{mildenhall2021nerf} treats input coordinates as continuous variables, offering flexible resolution for function outputs. 
Neural operators~\cite{anandkumar2019neural,li2021fourier,Kovachki2023,azizzadenesheli2024neural} and Galerkin transformers~\cite{cao2021choose} further generalize neural architectures to process sets of points as functional inputs, enabling function-space learning. 
Recently, a neural operator with localized integral and differential kernels has been introduced to improve its capability in capturing local features~\cite{liu-schiaffini2024neural}. To handle complex and variable geometries, the point cloud neural operator has been proposed, which adaptively processes point cloud datasets~\cite{zeng2025pointcloudneuraloperator}.

\noindent \textbf{Generative Models in Function Space.}
Early Neural Processes~\cite{garnelo2018neuralprocesses,pmlr-v80-garnelo18a} drew upon Gaussian processes~\cite{rasmussen2003gaussian}, and later methods such as GASP~\cite{pmlr-v151-dupont22a}, Functa~\cite{pmlr-v162-dupont22a}, and GANO~\cite{rahman2022generative} treat data as function evaluations to enable discretization-independent learning. Energy-based and diffusion models~\cite{pmlr-v206-lim23a,pmlr-v206-kerrigan23a,lim2025scorebaseddiffusionmodelsfunction,pidstrigach2023infinitedimensionaldiffusionmodels,franzese2024continuous}, along with flow-based approaches like FFM~\cite{pmlr-v238-kerrigan24a} and OpFlow~\cite{shi2024universal}, further extend these ideas. Ultimately, developing comprehensive generative models in function space requires defining suitable stochastic processes, score operators, and consistent neural mappings, along with specialized training methods for numerical stability—challenges that remain largely unresolved.
 
\noindent \textbf{Airfoil Design and Optimization.}
Airfoil design is essential for aircraft and wind turbines. Geometric parameterization supports efficient shape modeling and optimization. Techniques such as Free-Form Deformation (FFD)~\cite{farin2002curves} and NURBS~\cite{schoenberg1964spline} are widely used in CAD/CAE tools~\cite{dannenhoffer2024overview,hahn2010vehicle}, but their independent control points can cause instability, high dimensionality, and suboptimal solutions. Modal methods, including Proper Orthogonal Decomposition~\cite{berkooz1993proper}, global modal~\cite{bruls2007global}, and compact modal~\cite{li2021adjoint} approaches, reduce dimensionality by encoding global features, though they struggle with large or detailed shape changes. Class-Shape Transformation (CST)~\cite{kulfan2008universal} offers interpretable and differentiable shapes but lacks flexibility for large deformations and is sensitive to parameter choices. Therefore, a high-degree-of-freedom representation is needed for robust, nonlinear shape modeling. Unlike general computer vision tasks~\cite{su2020adapting,kirillov2023segment}, engineering AI models are domain-specific and often limited by sparse data. Generative models are advantageous here, as they can exploit unlabeled or untested samples. In airfoil design, VAEs, GANs, and diffusion models~\cite{chen2021bezierganautomaticgenerationsmooth,li2022machine,xie2024parametric,yangaobo,zhenweidiffAirfoil,liu2024afbench}, along with CFD-based mesh representations~\cite{weizhen2023,ZhenWeiDeepGeo2024}, have been used to map latent spaces to airfoil shapes. However, most existing methods model discrete airfoil points, limiting flexibility for downstream applications.

\section{FuncGenFoil: Function-Space Generative Model for Airfoils}

In contrast to existing airfoil generation methods, FuncGenFoil is constructed as a function-space generative model capable of producing airfoil geometries as continuous functions rather than discrete points, thereby leveraging the advantages of both parametric-model-based and discrete-point-based methods.
In this section, we detail the processes for airfoil generation and editing tasks, respectively.

\subsection{Airfoil Generation}

\noindent \textbf{Airfoil Parametrization.} 
Since an airfoil curve $(x,y)$ has circular topology, we introduce the variable $\alpha \in [0,1]$ as the domain of the function, denoting $y(\alpha) = f(x(\alpha))$ and $x(\alpha) = \frac{\cos(2\pi \alpha) + 1}{2}$,\footnote{For the remainder of this paper, we omit the explicit dependence on $\alpha$ for clarity and convenience.} as shown in Figure~\ref{fig:reparameterization}. The entire FuncGenFoil framework is essentially an ordinary differential equation (ODE)-based generative model that produces airfoil curve functions by solving an ODE along continuous time $t \in [0,1]$ as $u_1 = u_0 +\int_{t=0}^{t=1} v_t\,\mathrm{d}t$, where $u_1=y$, starting from some latent function $u_0$.

\begin{figure*}[t]
    \small
    \centering
    \includegraphics[width=\linewidth]{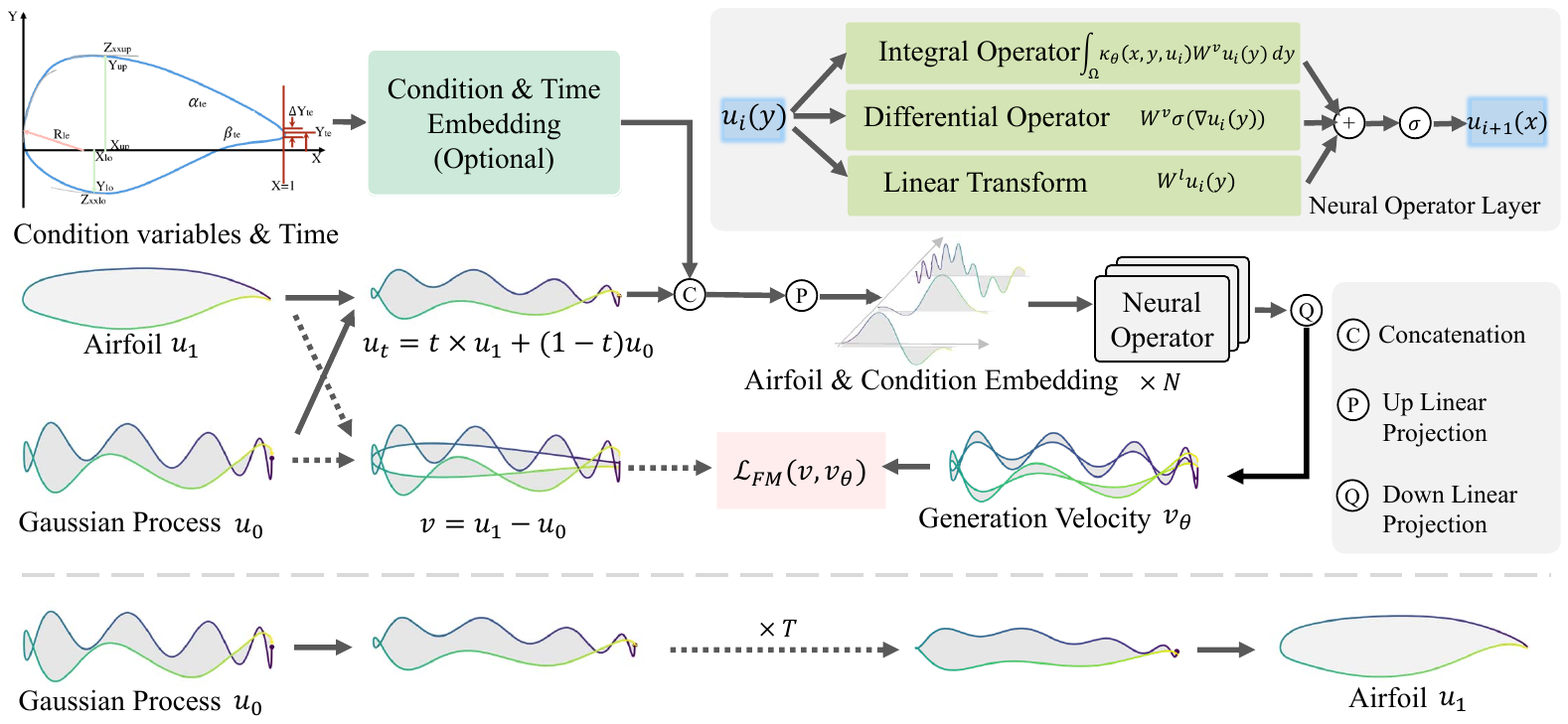}
    \caption{\textbf{Top:} Overview of \textbf{FuncGenFoil}'s neural network and training scheme. The model is a Fourier Neural Operator designed for point cloud data, although any other neural operator capable of general-purpose function-space approximation may be used.  
    The model takes as input a function $u_t$ (point cloud data at an arbitrary resolution $d$), optional design condition variables $c$, and the generation time $t$. It then processes this input function consistently and outputs the current velocity operator $v_{\theta}(u_t, c, t)$ for calculating the flow matching loss.
    \textbf{Bottom:} Inference with \textbf{FuncGenFoil} is conducted by first sampling a random latent function from a Gaussian process and then reconstructing the airfoil by solving an ODE.
    }
    \label{fig:velocity}
\end{figure*}

\noindent \textbf{Velocity Operator.} 
The velocity operator $v_t$ gradually transforms a latent function $u_0$ sampled from a stochastic process $\mathcal{P}$ into an airfoil function $u_1$ belonging to the target airfoil distribution $\mathcal{Q}$. The velocity operator is a key component, as it must handle function inputs consisting of point sets with arbitrary resolution or positions. Specifically, the operator takes as input the partially generated airfoil $u_t=[u_t(\alpha_0),u_t(\alpha_1),u_t(\alpha_2),\dots]$ and outputs a velocity function corresponding to these positions, $[v_t(\alpha_0),v_t(\alpha_1),v_t(\alpha_2),\dots]$.
We realize this by establishing a parameterized neural operator $v_{\theta}$ with model weights $\theta$, constructed as a Neural Operator model capable of processing function-space data at arbitrary resolutions in a single unified model, as shown in Figure~\ref{fig:velocity}.
Specifically, FuncGenFoil acts as a conditional continuous-time generative model, where the velocity operator $v_{\theta}(u_t, c, t)$ takes the noised airfoil $u_t$, optional conditioning variables $c$, and generation timestamp $t$ as inputs, and consistently outputs the deformation $v_t$ as a function.
We train $v_{\theta}(u_t, c, t)$ using Operator Flow Matching~\cite{shi2025stochasticprocesslearningoperator}.

\noindent \textbf{Training.} 
We train $v_{\theta}$ under the simplest denoising training process. Given an airfoil geometry $u_1$, we compute its corresponding noised sample at time $t$  as follows:
\begin{equation}\label{eq:path}
    u_t|_{u_0,u_1} = t\times u_1 + (1-t)\times u_0
\end{equation}
and then the ground-truth velocity $v_t(u_t)|_{u_0,u_1}$ given $u_0$ and $u_1$ can be computed as: \begin{equation}\label{eq:target_velocity}
    v_t(u_t)|_{u_0,u_1} = \frac{du_t}{dt} = u_1 - u_0.
\end{equation}
We can train $v_{\theta}$ by matching the velocity operator using Flow Matching loss:
\begin{equation}\label{eq:flow_matching}
    \mathcal{L}_{\mathrm{FM}}=E_{t\sim[0,1],u_t}\left[\|v_t(u_t)-v_{\theta}(u_t,c,t)\|^2\right],
\end{equation}
which according to Conditional Matching Theorem, has same gradients with Conditional Flow Matching loss:
\begin{equation}\label{eq:conditional_flow_matching}
    \mathcal{L}_{\mathrm{CFM}}=E_{t\sim[0,1],u_0,u_1}\left[\|v_t(u_t)|_{u_0,u_1}-v_{\theta}(u_t|_{u_0,u_1},c,t)\|^2\right],
\end{equation}
\begin{equation}\label{eq:conditional_matching_theorem}
    \nabla_{\theta}\mathcal{L}_{\mathrm{FM}}(\theta)=\nabla_{\theta}\mathcal{L}_{\mathrm{CFM}}(\theta).
\end{equation}
See more proofs in Chapter 4 of \textit{Flow Matching Guide and Code}~\citep{lipman2024flowmatchingguidecode}.

\noindent \textbf{Inference.}  Given a trained velocity operator $v_{\theta}$, the inference process, \emph{i.e.,} airfoil generation process, is equal to deriving airfoil geometry at time $t=1$, denoted as $u_1$, based on a latent coding $u_0$ sampled from the stochastic process $\mathcal{P}$. $\mathcal{P}$ is assumed as a Gaussian Process $\mathcal{GP}(0, K)$ in this work, where $K$ is a covariance kernel function. Therefore, $u_1$ could be derived by solving the generation ODE numerically (e.g., using the Euler method) as follows:
\begin{equation}
    u_1 = u_0 + \int_{t=0}^{t=1} v_{\theta}(u_t, c, t)\,\mathrm{d}t.
\end{equation}

Detailed implementations for training velocity operator and model inference are shown in Appendix~\ref{sec:appendix_algorithm}, Algorithm~\ref{alg:pretrain} and Algorithm~\ref{alg:generation}.

\subsection{Airfoil Editing}

\begin{figure}[t]
    \small
    \centering
      \includegraphics[width=\linewidth]{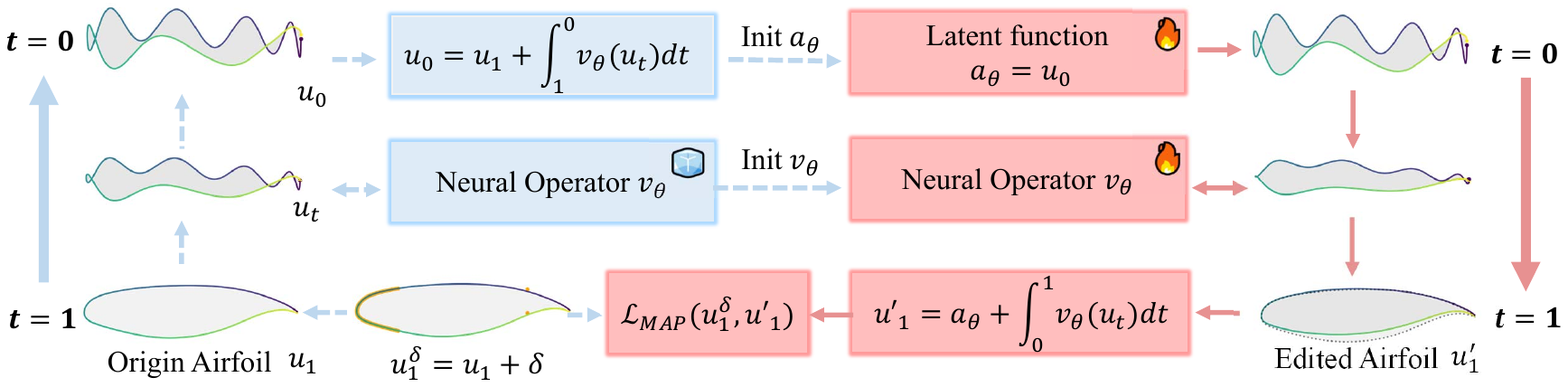}
      \caption{Airfoil editing by \textbf{FuncGenFoil}. Given a original airfoil $u_1$, an editing requirement $\delta$ and target airfoil $u_1^{\delta}$. We first infer its latent function $u_0$ reversely, and make it learnable as $a_{\theta}$. Then we sample a new airfoil $u_1^{'}$, and conduct a regression in function space via maximum a posteriori estimation $\mathcal{L}_{\mathrm{MAP}}$. After a few iterations of fine-tuning, we can generate edited airfoil $u_1^{\delta}$ with high accuracy.}
      \label{fig:editing}
      
\end{figure}

The airfoil editing task enables the user to modify parts of the geometry of a given airfoil $u_1$, effectively generating a new airfoil geometry $u_1^{'}$ while preserving the user-edited sections, denoted as $u_1^{\delta}$. 
We achieve this by finetuning the pretrained model via maximum a posteriori (MAP) estimation, $\max p(u_1^{'} \mid u_1^{\delta})$ where $p(u_1^{'} \mid u_1^{\delta})$ is a probabilistic model that constrains the optimized airfoil $u_1^{'}$ fulfilling the editing requirements and following the generation prior.

To achieve the constraint probability model, we disentangled it with Bayes' Rule:
\begin{equation}
\label{eq:map}
    \max_{u_1^{'}} p(u_1^{'} \mid u_1^{\delta})
    \;=\;
    \frac{p(u_1^{\delta} \mid u_1^{'})\cdot p(u_1^{'})}{p(u_1^{\delta})}
    \Rightarrow
    \max_{u_1^{'}} \log{p(u_1^{\delta}|u_1^{'})} + \log{p(u_1^{'})} - \log{p(u_1^{\delta})},
\end{equation}

where $p(u_1^{\delta} \mid u_1^{'})$ is a Gaussian measure, so its log term becomes a Mean Square Error (MSE) between $u_1^{\delta}$ and $u_1^{'}$, constraining the user edited parts in $u_1^{'}$ keeping consistent with $u_1^{\delta}$. We can sample $u_1^{'}$ through solving the velocity operator using Neural ODE method\cite{Chen2018} and calculate $\log p(u_1^{'})$ at the same time using FFJORD method \cite{Grathwohl2018} with Hutchinson trace estimator \cite{Hutchinson1989}.
$p(u_1^{'})$ is the prior supported by the trained generative model. 
$p(u_1^{\delta})$ is the marginal likelihood, which does not depend on $u_1^{'}$. The final optimization target could be written as:

\begin{equation} \small
\label{eq:edit_loss}
\begin{aligned}
    \max_{u_1^{'}} \frac{1}{2\sigma^2} \sum_{i\in\Delta} (u_1^{\delta,i} - u_1^{',i})^2 + \log{p(u_1^{'})},
\end{aligned}
\end{equation}
where  $\sigma$ is noise scale for editing, $\Delta$ denotes point indices in edited part $u_1^{\delta}$.

For more realistic generation results, the optimization does not directly adjust $u_1^{'}$; instead, we optimize $u_1^{'}$ indirectly by fine-tuning the entire generative model for a few iterations. The fine-tuning process is illustrated in Figure~\ref{fig:editing}, while its details are provided in Alg.~\ref{alg:editing} in the Appendix.
Specifically, we first initialize $u'_1$ as a resample data of $u_1$, by extracting its latent code, denoted as $\alpha_\theta$, via the inverse of our generative model and re-generating $u_1^{'}$ from this code. Then we treat Equation~\ref{eq:edit_loss} as the loss function to train $\theta$ of the velocity operator and $\alpha_\theta$ simultaneously. After the model training, the new $u_1^{'}$ is the new edited generation results.

\section{Experiments}
\subsection{Experimental Settings}

\noindent\textbf{Tasks.}
We evaluate two tasks within the airfoil inverse design problem: \textit{conditional airfoil generation} and \textit{freestyle airfoil editing}.
In the conditional generation task, the model is provided with a set of 11 geometric parameters describing the airfoil geometry. Detailed parameter definitions are provided in Table~\ref{tab:geoparams} in the Appendix. The model must generate airfoils that satisfy these geometric constraints.
In the freestyle editing task, the model receives an original airfoil and a target modification, such as adjusting the position of a specific point on the airfoil. \emph{The selected point can be located anywhere on the airfoil surface.} The model must generate an airfoil reflecting the specified modification.

\noindent\textbf{Metrics.} We adopt the metrics introduced in AFBench~\cite{liu2024afbench} to evaluate the generated airfoils:

\underline{\textit{Label Error}} measures the difference between the PARSEC parameters of the generated or edited airfoil and the intended target parameters, calculated as $\sigma_i = \left| \hat{p}_i - p_i \right|$,
where $\sigma_i$ is the label error for the $i$-th parameter, $\hat{p}_i$ is the $i$-th geometric parameter of the generated airfoil, and $p_i$ is the corresponding target parameter. Smaller values indicate better alignment with target parameters. To summarize all $11$ label errors, we present both the arithmetic mean $\bar{\sigma}_a$ and the geometric mean $\bar{\sigma}_g$ for absolute and relative average errors.

\underline{\textit{Diversity}} quantifies the variety among generated airfoils, calculated as $D = \frac{1}{n} \sum_{i=1}^{n} \log \det(L_{S_i})$,
where $n$ is the number of subsets, and $L_{S_i}$ is the similarity matrix of the $i$-th subset, computed based on Euclidean distances between airfoils within the subset. Higher values indicate greater diversity among generated airfoils.

\underline{\textit{Smoothness}} measures the geometric smoothness of generated airfoils, calculated as:
\begin{equation} \small
    M = \sum_{n=1}^{N} \text{Distance}(P_n, P_{n-1} P_{n+1}),
\end{equation}
where $P_n$ is the $n$-th point, and $P_{n-1} P_{n+1}$ is the line segment connecting its adjacent points. The function $\text{Distance}(P_n, P_{n-1} P_{n+1})$ computes the perpendicular distance from $P_n$ to this line segment. Smaller values indicate better geometric quality.

\noindent\textbf{Datasets.} To benchmark our method, we conduct experiments on three datasets: UIUC~\cite{selig1996uiuc}, Supercritical Airfoil (Super), and AF-200K.
UIUC contains 1,600 designed airfoil geometries. Super focuses on supercritical airfoils and includes approximately 20,000 airfoil samples. AF-200K consists of 200,000 highly diversified airfoil samples.

\noindent\textbf{Baselines.} We compare FuncGenFoil with baseline models proposed in AFBench~\cite{liu2024afbench}, specifically the conditional VAE (CVAE), conditional GAN (CGAN), the modified VAE with PARSEC parameters and control keypoints (PK-VAE), as well as PK-GAN, PKVAE-GAN, the U-Net-based PK-DIFF, and the transformer-based PK-DIT.

\begin{table*}[t]
    \centering
    \renewcommand{\arraystretch}{1.2}
    \setlength{\tabcolsep}{4pt}
    \caption{Quantitative comparison between FuncGenFoil and baseline methods across different datasets for the conditional generation task. Label error, diversity, and smoothness of generated airfoils are reported.}
    \label{tab:airfoil_results}
    \resizebox{\textwidth}{!}{
    \begin{tabular}{llccccccccccccccc}
        \toprule
        \multirow{2}{*}{\textbf{Method}} & \multirow{2}{*}{\textbf{Dataset}} & \multicolumn{13}{c}{\textbf{Label Error \boldmath$\downarrow (10^{-3})$}} & \multirow{2}{*}{\boldmath$\mathcal{D} \uparrow$} & \multirow{2}{*}{\boldmath$\mathcal{M} \downarrow (10^{-2})$} \\ \cline{3-15}
 &  & $\sigma_1$ & $\sigma_2$ & $\sigma_3$ & $\sigma_4$ & $\sigma_5$ & $\sigma_6$ & $\sigma_7$ & $\sigma_8$ & $\sigma_9$ & $\sigma_{10}$ & $\sigma_{11}$ & $\bar{\sigma}_a$ & $\bar{\sigma}_g$ &  \\ \hline
        CVAE     & AF-200K & 72.9  & 52.5  & 35.2  & 15900  & 99  & 95  & 29000  & 19.1  & 15.3  & 46  & 104  & 4131 & 149.2  & -155.4  & 7.09 \\
        CGAN     & AF-200K & 107  & 85.0  & 54.4  & 23200  & 143 & 137 & 59600 & 25.3  & 22.3  & 53  & 129  & 7596   &  217.6 & -120.5  & 7.31 \\
        PK-VAE   & AF-200K & 63.0  & 47.9  & 31.3  & 8620   & 66  & 64  & 17100  & 13.5  & 9.3  & 33  & 78   & 2375  &  106.0 & -150.1  & 5.93 \\
        PK-GAN   & AF-200K & 81.8  & 63.0  & 47.0  & 21030  & 120 & 117 & 32470  & 22.5  & 19.6  & 50  & 122  & 4922 & 179.5  & -112.3  & 3.98 \\
        PKVAE-GAN& AF-200K & 56.8  & 31.7  & 31.0  & 5650   & 46  & 43  & 12000  & 9.1  & 5.1  & 28  & 63   & 1633   & 77.6 & -129.6  & 2.89 \\
        PK-DIT   & AF-200K & 11.2 & 32.3 & 15.4 & 1050 & \textbf{13} & 11.5 & 9790 & 0.5 & \textbf{0.5} & \textbf{23} & \textbf{24} & 997 & 23.5 & -93.2 & \textbf{1.04} \\
        \rowcolor{gray!20} FuncGenFoil & AF-200K & \textbf{1.84} & \textbf{17.4} & \textbf{0.56} & \textbf{721} & 47.7 & \textbf{0.98} & \textbf{1676} & \textbf{0.45} & 0.65 & 160 & 174 & \textbf{255}  &  \textbf{14.9}  & \textbf{-71.6} & 1.41 \\
        \midrule
        PK-VAE & UIUC & 80.7 & 20.9 & 12.2 & 12843 & 36.9 & 14.0 & 37263 & 1.7 & 1.9 & 94.8 & 109.9 & 4589 & 69.1 & -93.5 & 7.29 \\
        PK-DIT & UIUC & 63.8 & 51.4 & 33.6 & 11830 & 87 & 84.9 & 25700 & 16.9 & 11.9 & \textbf{36} & \textbf{98} & 3456 & 129.8 & -141.7 & 6.03 \\
        \rowcolor{gray!20} FuncGenFoil & UIUC & \textbf{11.6} & \textbf{11.8} & \textbf{0.38} & \textbf{698} & \textbf{23.5} & \textbf{0.56} & \textbf{12339} & \textbf{0.27} & \textbf{0.44} & 110.5 & 115.0 & \textbf{1210} & \textbf{15.1} & \textbf{-91.1} & \textbf{1.33} \\
        \midrule
        PK-VAE & Super & 10.8 & 17.5 & 2.3 & 1735.6 & 12.1 & 3.2 & 8131 & 3.5 & 1.4 & 98.3 & 80.6 & 918 & 28.4 & -122.8 & 1.38 \\
        PK-DIT & Super & 52.0 & 35.0 & 24.0 & 3010 & 29 & 33.2 & 10500 & 8.3 & 2.6 & \textbf{27} & \textbf{33} & 1250 & 58.1 & -123.4 & 1.97 \\
        \rowcolor{gray!20} FuncGenFoil & Super & \textbf{0.71} & \textbf{8.23} & \textbf{0.13} & \textbf{201.3} & \textbf{4.72} & \textbf{0.12} & \textbf{174.2} & \textbf{0.09} & \textbf{0.14} & 34.2 & 36.7 & \textbf{41.9}   & \textbf{3.08}  & \textbf{-103.9}  & \textbf{1.01} \\
        \bottomrule
    \end{tabular}
    }
    \label{tab:cond_gen}
\end{table*}

\begin{table*}[ht]
    \centering
    \caption{Quantitative evaluation of the FuncGenFoil model across different sampling super-resolutions for the conditional generation task. The training resolution is $257$.}
    \label{tab:airfoil_results_super_resolution}
    \resizebox{\textwidth}{!}{
    \begin{tabular}{ccccccccccccccccc}
        \toprule
        \multirow{2}{*}{\textbf{Dataset}}
        & \multirow{2}{*}{\textbf{Resolution}}
        & \multicolumn{13}{c}{\textbf{Label Error \boldmath$\downarrow (10^{-3})$}} 
        & \multirow{2}{*}{\boldmath$\mathcal{D} \uparrow$} 
        & \multirow{2}{*}{\boldmath$\mathcal{M} \downarrow (10^{-2})$} 
        \\ \cline{3-15}
 & & $\sigma_1$ & $\sigma_2$ & $\sigma_3$ & $\sigma_4$ & $\sigma_5$ & $\sigma_6$ & $\sigma_7$ & $\sigma_8$ & $\sigma_9$ & $\sigma_{10}$ & $\sigma_{11}$ & $\bar{\sigma}_a$ & $\bar{\sigma}_g$ &  \\ \hline
        \multirow{3}{*}{Super} 
        & 257 & 0.71 & 8.23 & 0.13 & 201.3 & 4.72 & 0.12 & 174.2 & 0.09 & 0.14 & 34.2 & 36.7 & 41.9 & 3.08 & -103.9 & 1.01 \\
        & 513 & 0.71 & 8.19 & 0.12 & 200.8 & 4.81 & 0.12 & 174.7 & 0.09 & 0.15 & 48.2 & 47.8 & 44.1 & 3.26 & -97.5 & 0.52 \\
        & 1025 & 0.71 & 8.29 & 0.12 & 202.9 & 4.93 & 0.12 & 175.4 & 0.09 & 0.15 & 61.4 & 57.8 & 46.5 & 3.44 & -91.1 & 0.37 \\
        \bottomrule
    \end{tabular}
    }
\end{table*}

\noindent\textbf{Implementation Details.} On the AF-200K dataset, we trained for 2 million iterations with a batch size of 2,048 using 8 NVIDIA 4090 GPUs. On the Supercritical Airfoil and the UIUC dataset, we trained for 1 million iterations with a batch size of 1,024 on 4 single NVIDIA 4090 GPUs. We use Gaussian processes with a Matérn kernel function ($\nu=2.5, l=0.03$) as the prior. Other training hyperparameters are detailed in Appendix~\ref{app:hyper}.

\subsection{Main Results}
\noindent\textbf{Conditional Airfoil Generation.}
As shown in Table~\ref{tab:cond_gen}, FuncGenFoil outperforms the strongest baseline method (PK-DIT) across all metrics.
For \textit{label error}, FuncGenFoil achieves reductions in arithmetic mean label error of \textbf{74.4\%} on AF-200K, \textbf{65.0\%} on UIUC, and \textbf{96.6\%} on Super. More importantly, it also reduces the geometric mean label error by \textbf{36.6\%} on AF-200K, \textbf{88.4\%} on UIUC, and \textbf{94.7\%} on Super.
This substantial decrease in label error underscores the effectiveness of our approach for generating airfoils that more precisely adhere to target geometric parameters.
In terms of \textit{diversity}, FuncGenFoil demonstrates notable improvements, surpassing the best baseline methods by \textbf{23.2\%}, \textbf{35.7\%}, and \textbf{15.8\%} on AF-200K, UIUC, and Super, respectively.
This highlights the model's superior capability in capturing and generating a broader spectrum of valid airfoil designs.
Additionally, the generated airfoils exhibit enhanced surface \textit{smoothness}, as evidenced by reductions of \textbf{4.70} and \textbf{0.96} in smoothness values ($10^{-2}$) on the UIUC and Super datasets, respectively.
This improvement is particularly crucial for aerodynamic performance, as smoother airfoil surfaces contribute to reduced drag and improved flow characteristics. 
The limited coverage of the training dataset constrains the model's effectiveness to a specific operational range, which is detailed in Table~\ref{tab:effective_range}.
A performance comparison between FuncGenFoil and classical methods is provided in Appendix~\ref{app:tradition_method}.

\begin{figure}[t]
    \small
    \centering
    \includegraphics[width=0.9\linewidth]{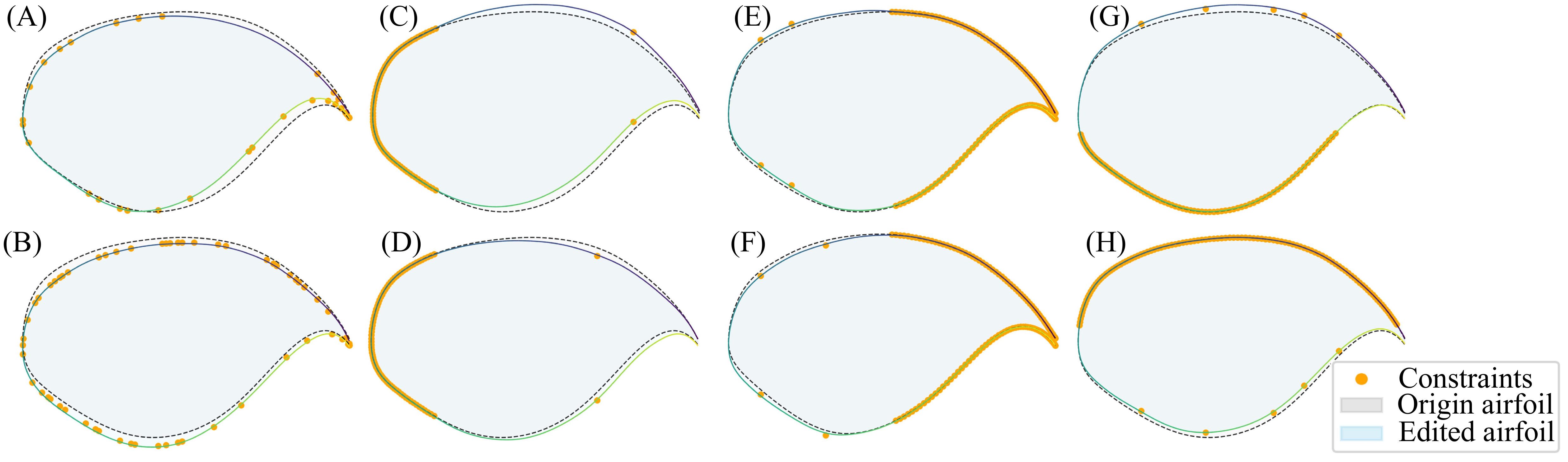}
    \caption{Examples of eight different FuncGenFoil instances performing airfoil editing over 20 training iterations. The orange points and sections represent the editing requirements as constraints; the gray region is the original airfoil, while the blue region shows the generated edited airfoil. The results demonstrate that the generated airfoil quickly adapts to the editing requirements within a few iterations, achieving a natural and smooth function regression. The editing scheme can be completely customized according to the user's preference.}
    \label{fig:airfoil_edit_finetuning}
\end{figure}

\begin{wraptable}{r}{0.5\textwidth}
\centering
\caption{Quantitative evaluation of the airfoil editing task across different editing scales. Mean squared error (MSE) between generated and target airfoils and smoothness of generated airfoils are reported.}
\scriptsize 
    \begin{tabular}{cccccc}
        \toprule
        \textbf{Dataset} & \textbf{Edit Scale} & \textbf{MSE} \boldmath$\downarrow ( 10^{-7})$ & {\boldmath$\mathcal{M} \downarrow (10^{-2})$} \\
        \hline
        \multirow{6}{*}{Super} & 0.0001 & 2.41 & 1.16 \\
                               & 0.0002 & 2.45 & 1.15 \\
                               & 0.0004 & 2.75 & 1.15 \\
                               & 0.0008 & 4.32 & 1.26 \\
                               & 0.0016 & 15.5 & 1.35 \\
                               & 0.0032 & 61.7 & 1.49 \\
        \bottomrule
    \end{tabular}
 \label{tab:edit}
\end{wraptable}
\noindent\textbf{Freestyle Airfoil Editing.} Table~\ref{tab:edit} shows average performance over 300 editing cases using FuncGenFoil. In each case, the model adjusts 2 to 4 randomly selected positions on the airfoil surface to target locations over 10 fine-tuning steps, with editing scales ranging from $1 \times 10^{-4}$ to $3.2 \times 10^{-3}$. Results demonstrate that FuncGenFoil achieves accurate airfoil editing with minimal errors (less than \boldmath{$2.75 \times 10^{-7}$} MSE) and high surface smoothness (less than  $1.16 \times 10^{-2}$ smoothness value) after only a few fine-tuning steps if the edit scale is less than $4 \times 10^{-4}$. The edit error increases at an accelerating rate with larger edits but remains relatively low overall.
Figure~\ref{fig:airfoil_edit_finetuning} illustrates example editing requirements and corresponding generated airfoils during the fine-tuning stage, demonstrating that our model effectively completes freestyle editing tasks by generating accurate airfoils according to user-specified edits.

\noindent\textbf{Any-Resolution Airfoil Generation.}
One of the key advantages of FuncGenFoil is its ability to generate airfoils at any resolution while maintaining high generation quality. This is achieved by representing airfoils as functions and learning resolution-independent function transformations, enabling flexible and consistent airfoil generation across different scales.
To evaluate high-resolution generation, we use the model trained on the Supercritical Airfoil dataset at a resolution of 257 and sample new airfoils at resolutions of 513 and 1025. We then assess these higher-resolution airfoils using the same metrics as in the conditional generation task. The results are presented in Table~\ref{tab:airfoil_results_super_resolution}.
We observe consistent generation quality at 2$\times$ and even 4$\times$ the training resolution, with a maximum increase of \textbf{4.6$\times10^{-3}$} in average label error, a decrease of $12.8$ in diversity, and an increase of \textbf{0.64$\times10^{-2}$} in smoothness value.

\begin{figure}[t]
    \centering
    \includegraphics[width=\linewidth]{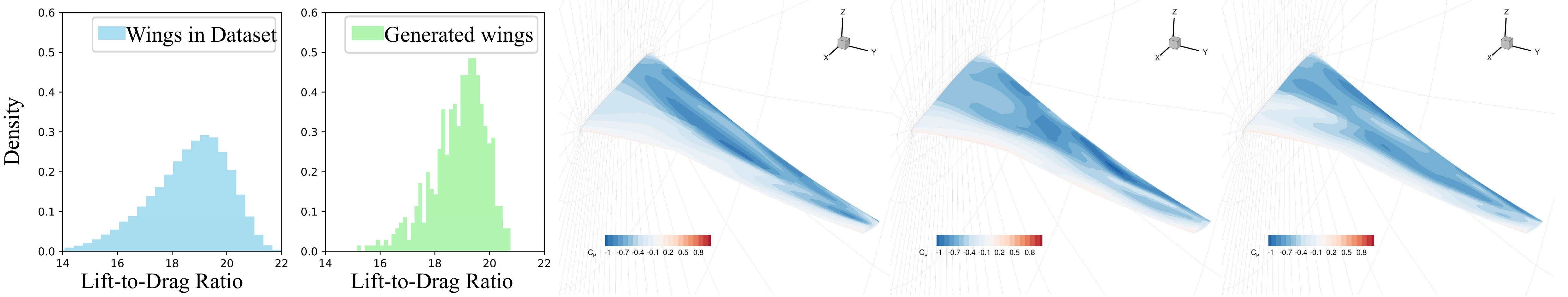}
    \caption{\textbf{Left}: Comparison of the lift-to-drag ratio histograms for CRM wings in the original dataset and the samples generated by our FuncGenFoil model. \textbf{Right}: Visualization of the pressure coefficient for the generated CRM wings, obtained through aerodynamic simulation.}
    \label{fig:crm_lift_drag_cp}
\end{figure}

\noindent\textbf{Aerodynamic Simulation.}
To further assess the physical properties of the generated airfoils and validate the effectiveness of our method, we analyze their aerodynamic performance using the NASA Common Research Model (CRM)~\footnote{\url{https://commonresearchmodel.larc.nasa.gov/}} and perform Reynolds-Averaged Navier-Stokes (RANS) computational fluid dynamics (CFD) simulations on the generated samples.
The CRM dataset contains 135,000 CRM wing geometries along with their corresponding aerodynamic performance, computed using the RANS CFD solver ADflow~\cite{Mader2020a}.
We pretrain the FuncGenFoil model on all 135,000 CRM wing geometries and generate 500 new CRM wing geometries for RANS CFD evaluation. We analyze the lift-to-drag ratio ($L/D$) of these newly generated geometries and compare them with the original dataset, as shown in Figure~\ref{fig:crm_lift_drag_cp}. 
Our results indicate that the $L/D$ distribution of the generated samples closely aligns with that of the original dataset, with the highest density occurring around $L/D = 19$, demonstrating the model's ability to learn and generate physically plausible wing geometries. Additionally, we visualize the coefficient of pressure contours for selected CFD cases in Figure~\ref{fig:crm_lift_drag_cp}. These visualizations confirm that FuncGenFoil can generate airfoils with diverse aerodynamic performance characteristics.

\subsection{Ablation Study}

\begin{figure}[t]
    \centering
    \includegraphics[width=\linewidth]{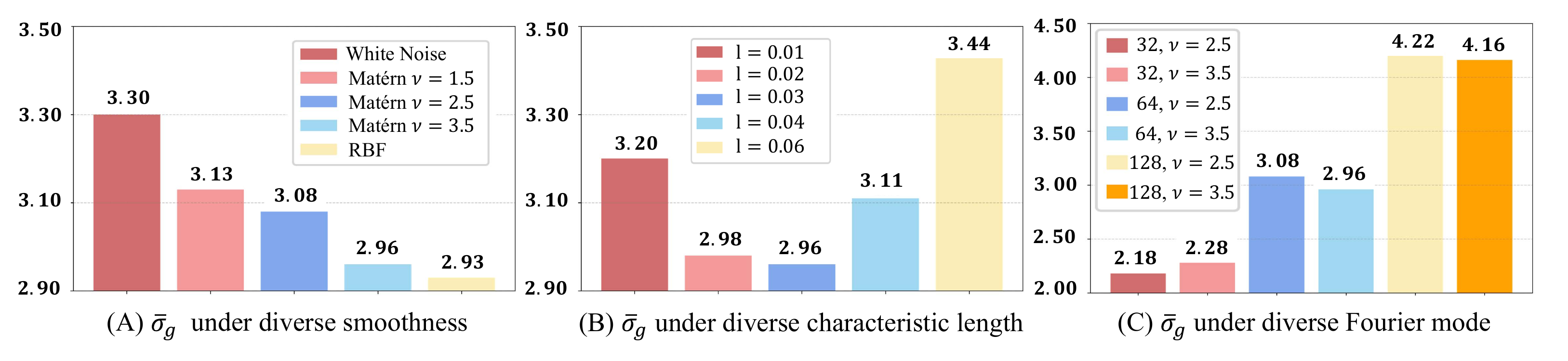}
    \caption{\textbf{(A)}: Comparison of the geometric mean error $\bar{\sigma}_g$ using kernels of different smoothness but same characteristic length $l=0.03$. \textbf{(B)}: Comparison of the geometric mean error $\bar{\sigma}_g$ using Matérn kernel $\nu=3.5$ with different characteristic length $l$. \textbf{(C)}: Comparison of the geometric mean error $\bar{\sigma}_g$ using different Fourier mode in FNO using Matérn kernel of same $l=0.03$.}
    \label{fig:ablation_kernel}
\end{figure}

\textbf{Kernels of Gaussian Process.}
The type of Gaussian process kernel used as a prior in the model influences the function space into which the data is diffused. We perform ablation experiments across several kernel types, including the white noise kernel, the Matérn kernel with smoothness parameter $\nu \in [1.5, 2.5, 3.5]$, and the radial basis function (RBF) kernel. We measure the geometric mean $\bar{\sigma}_g$ under different smoothness, characteristic length, and Fourier mode settings, as shown in Figure~\ref{fig:ablation_kernel}. We observe that FunGenfoil benefits from employing smoother kernels, which enhance both the smoothness and accuracy of the generated airfoils. This effect arises because the parameter $\nu$ primarily controls the smoothness of the function space, and the RBF kernel corresponds to the case $\nu \rightarrow \infty$. Additionally, we find that there exists an optimal characteristic length at $l = 0.03$ for the kernel. In our experiments, FunGenfoil does not exhibit improved performance with an increased number of Fourier modes. This phenomenon may be due to the predominantly low-frequency characteristics of supercritical airfoils, such that adding higher modes in the FNO layer introduces unnecessary model complexity. More detailed experimental results are provided in Table~\ref{tab:ablation_matern} in the Appendix.

\begin{wraptable}{r}{0.4\textwidth}
\centering
\caption{Ablation study on the effect of different latent variable initialization methods for the airfoil editing task.}
\scriptsize 
\begin{tabular}{ccc}
\toprule
\multirow{2}{*}{\textbf{Edit scale}} & \multicolumn{2}{c}{\textbf{MSE}$\mathcal\downarrow (10^{-7})$}            \\ \cline{2-3} 
                            & \textbf{w/} ODE inv. & \textbf{w/o} ODE inv. \\ \midrule
0.0001                      & 2.42 \tiny($\downarrow$83.88)           & 86.3             \\
0.0002                      & 2.46 \tiny($\downarrow$84.14)           & 86.6             \\
0.0004                      & 2.75 \tiny($\downarrow$84.85)           & 87.6             \\
\bottomrule
\end{tabular}
\label{tab:ablate_initialize}
\end{wraptable}
\textbf{Editing with Prior via ODE Inversion.} In Table~\ref{tab:ablate_initialize}, we compare MSE of constraint condition calculations in the airfoil editing task using two initialization schemes: (1) using the latent variables obtained through ODE inversion of the original airfoil as the prior, and (2) using a zero prior $u_0=0$ without ODE inversion. Both schemes are fine-tuned with the same number of iterations. The ODE inversion-based prior significantly reduces the MSE compared to the zero prior, with a maximum reduction of \textbf{84.85}.
This underscores the importance of incorporating original airfoil information during the initialization process.

\textbf{ODE Numerical Method.}
Model inference involves solving an ordinary differential equation (ODE), and the choice of numerical integration scheme affects its performance. Using the same velocity operator $v_{\theta}$ trained on the supercritical airfoil dataset and sampling from a Gaussian prior with a Matérn kernel ($\nu=2.5$, $l=0.03$), we evaluate FuncGenFoil across various time step sizes and ODE solvers (Euler, midpoint, and fourth-order Runge-Kutta (RK4) methods). Results in Table~\ref{tab:ablate_ode_solver} show that FuncGenFoil benefits from larger time steps using Euler method. However, RK4 may degrade performance at excessively small step sizes, potentially because the model enters regions where the velocity operator is inadequately trained.
\begin{table*}[t]
    \centering
    \renewcommand{\arraystretch}{1.2}
    \setlength{\tabcolsep}{4pt}
    \caption{Quantitative experiments on the impact of ODE solver and time step on sampling quality.}
    \label{tab:ablate_ode_solver}
    \resizebox{\textwidth}{!}{
    \begin{tabular}{ccccccccccccccccc}
        \toprule
        \multirow{2}{*}{\textbf{ODE Solver}} & 
        \multirow{2}{*}{\textbf{Time Steps}} & 
        \multicolumn{13}{c}{\textbf{Label Error \boldmath$\downarrow (10^{-3})$}} &
        \multirow{2}{*}{\boldmath$\mathcal{D} \uparrow$} 
        & \multirow{2}{*}{\boldmath$\mathcal{M} \downarrow (10^{-2})$} \\ 
        \cline{3-15}
        & & 
        $\sigma_1$ & $\sigma_2$ & $\sigma_3$ & $\sigma_4$ & $\sigma_5$ & $\sigma_6$ & $\sigma_7$ & $\sigma_8$ & $\sigma_9$ & $\sigma_{10}$ & $\sigma_{11}$ & $\bar{\sigma}_a$ & $\bar{\sigma}_g$ 
        &  & \\ 
        \midrule
        \multirow{3}{*}{Euler} 
        & $10$ & 
        0.71 & 8.23 & 0.13 & 201.3 & 4.72 & 0.12 & 174.2 & 0.09 & 0.14 & 34.2 & 36.7 & 41.9 & 3.08 & -103.9 & 1.01 \\
        & $50$ & 
        0.62 & 3.61 & 0.07 & 92.5 & 2.08 & 0.06 & 96.8 & 0.06 & 0.07 & 14.5 & 15.7 & 20.6 & 1.60 & -107.6 & 0.99 \\
        & $100$ & 
        0.67 & 3.13 & 0.09 & 83.9 & 1.84 & 0.08 & 90.6 & 0.07 & 0.06 & 12.3 & 13.1 & 18.7 & 1.55 & -106.4 & 0.99 \\
        \midrule
        \multirow{3}{*}{Midpoint} 
        &  $10$ &
        0.68 & 5.29 & 0.08 & 161.3 & 3.12 & 0.08 & 178.9 & 0.06 & 0.09 & 21.4 & 22.4 & 35.8 & 2.17 & -105.1 & 1.00 \\
        &  $50$ &
        0.75 & 3.82 & 0.12 & 137.9 & 2.34 & 0.10 & 153.4 & 0.08 & 0.06 & 14.1 & 14.9 & 29.8 & 1.97 & -103.3 & 0.99 \\
        &  $100$ &
        0.82 & 3.94 & 0.14 & 125.7 & 2.22 & 0.13 & 137.4 & 0.08 & 0.07 & 14.4 & 14.2 & 27.2 & 2.06 & -101.6 & 0.99 \\
        \midrule
        \multirow{3}{*}{RK4} 
        &  $10$ &
        15.9 & 46.3 & 1.21 & 4406 & 25.7 & 1.24 & 4795 & 1.17 & 1.68 & 333 & 396 & 911 & 36.6 & -67.5 & 3.05 \\
        &  $50$ &
        0.92 & 7.63 & 0.24 & 482.7 & 5.55 & 0.22 & 511.8 & 0.16 & 0.19 & 39.3 & 41.3 & 99.1 & 4.68 & -92.9 & 1.07 \\
        &  $100$ &
        0.82 & 5.65 & 0.18 & 160.6 & 3.21 & 0.17 & 170.8 & 0.10 & 0.12 & 22.0 & 22.8 & 35.1 & 2.78 & -96.9 & 1.00 \\
        \bottomrule
    \end{tabular}
    }
    \label{tab:super_res}
\end{table*}

\section{Conclusion and Limitations}

In this work, we tackle the critical challenge of generating high-fidelity airfoil geometries that effectively balance expressiveness, smoothness, and resolution flexibility. We introduce \textbf{FuncGenFoil}, a function-space generative model leveraging neural operators and flow matching, which represents airfoils as continuous, smooth geometries without resolution constraints while preserving the expressiveness of data-driven methods. Comprehensive experimental results show that FuncGenFoil outperforms state-of-the-art techniques in terms of label error, diversity, and smoothness, highlighting its potential for high-fidelity airfoil design. Furthermore, the generated wing geometries have been validated through aerodynamic simulations. This work paves the way for more efficient, scalable, and versatile airfoil generation, with significant applications in aerodynamic shape optimization for aircraft manufacturing.

As for limitations, although we have taken a step toward general object shape modeling in function space, the focus of this paper is primarily on airfoils or aircraft wings, which have relatively simple geometries and smooth surfaces yet are highly significant for aerodynamic performance. Thus, our current approach has limitations regarding the scope of geometry. In future studies, if we aim to extend our method to modeling entire aircraft or dealing with objects of general shapes, substantial theoretical work and experimental analysis will be required, especially when a suitable coordinate system is unavailable for describing complex, irregular, or non-smooth geometries. This direction is part of our ongoing efforts.

\section*{Acknowledgements}
\textbf{Funding:} This work was supported by the Shanghai Artificial Intelligence Laboratory, the JC STEM Lab of AI for Science and Engineering, funded by The Hong Kong Jockey Club Charities Trust, the Research Grants Council of Hong Kong (Project No. CUHK14213224), the Natural Science Foundation of China (No. U23A2069), the AI for Science Seed Program of Shanghai Jiao Tong University (Project No. 2025AI4S-HY02), and the AI for Science Program, Shanghai Municipal Commission of Economy and Informatization (No. 2025-GZL-RGZN-BTBX-01010).

\textbf{Support and Collaboration:} We thank the team at Shanghai Aircraft Design and Research Institute and Shanghai Jiao Tong University for their valuable discussions and continued support.

\bibliographystyle{plain}
\bibliography{neurips_2025}

\newpage


\appendix

\newpage
\section{Problem Setting for Airfoil Generation}

\subsection{Airfoil Parametrization}

\begin{figure}[t]
\small
\centering
\includegraphics[width=\linewidth]{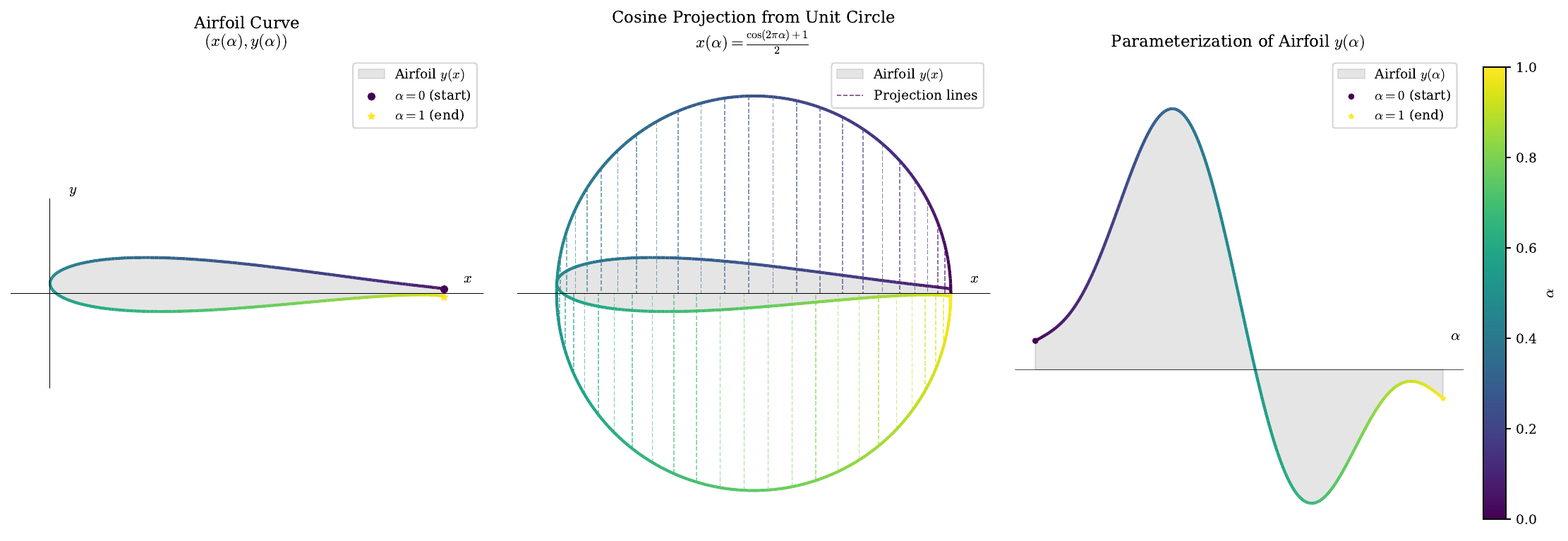}
\caption{
Illustration of the airfoil curve parameterization employed in FuncGenFoil.}
\label{fig:reparameterization}
\end{figure}

In this paper, we perform a circular parameterization of the airfoil curves, as illustrated in Figure~\ref{fig:reparameterization}. Specifically, we project a unit circle onto the $x$-axis and adopt the angle $\alpha$ as the function domain. The parameterized airfoil function begins at the trailing edge on the suction side, where $\alpha=0$, and ends at the trailing edge on the pressure side, where $\alpha=1$. In this manner, we transform the airfoil data from a circular topology into a well-defined functional form suitable for reconstruction by generative models.

\subsection{Design Parameters of Airfoil Geometry}

\label{app:groparams}
\begin{table}
  \centering
  \begin{subtable}[b]{0.5\textwidth}
    \centering
    \begin{tabular}{ccc}
      \toprule
      \textbf{Index} & Symbol & Geometric Meaning \\
      \midrule
      1  & $\mathrm{R_{le}}$   & leading edge radius \\
      2  & $\mathrm{X_{up}}$   & upper crest position x \\
      3  & $\mathrm{Y_{up}}$   & upper crest position y \\
      4  & $\mathrm{Z_{xxup}}$ & upper crest curvature \\
      5  & $\mathrm{X_{lo}}$   & lower crest position x \\
      6  & $\mathrm{Y_{lo}}$   & lower crest position y \\
      7  & $\mathrm{Z_{xxlo}}$ & lower crest curvature \\
      8  & $\mathrm{Y_{te}}$   & trailing edge position \\
      9  & $\Delta \mathrm{Y_{te}}$ & trailing thickness \\
      10 & $\alpha_{\mathrm{te}}$   & trailing edge angle up \\
      11 & $\beta_{\mathrm{te}}$    & trailing edge angle down \\
      \bottomrule
    \end{tabular}
    \caption{PARSEC parameters and its geometric meaning.}
    \label{tab:geoparams_a}
  \end{subtable}%
  \quad
  \begin{subtable}[b]{0.45\textwidth}
    \centering
    \includegraphics[width=0.8\linewidth]{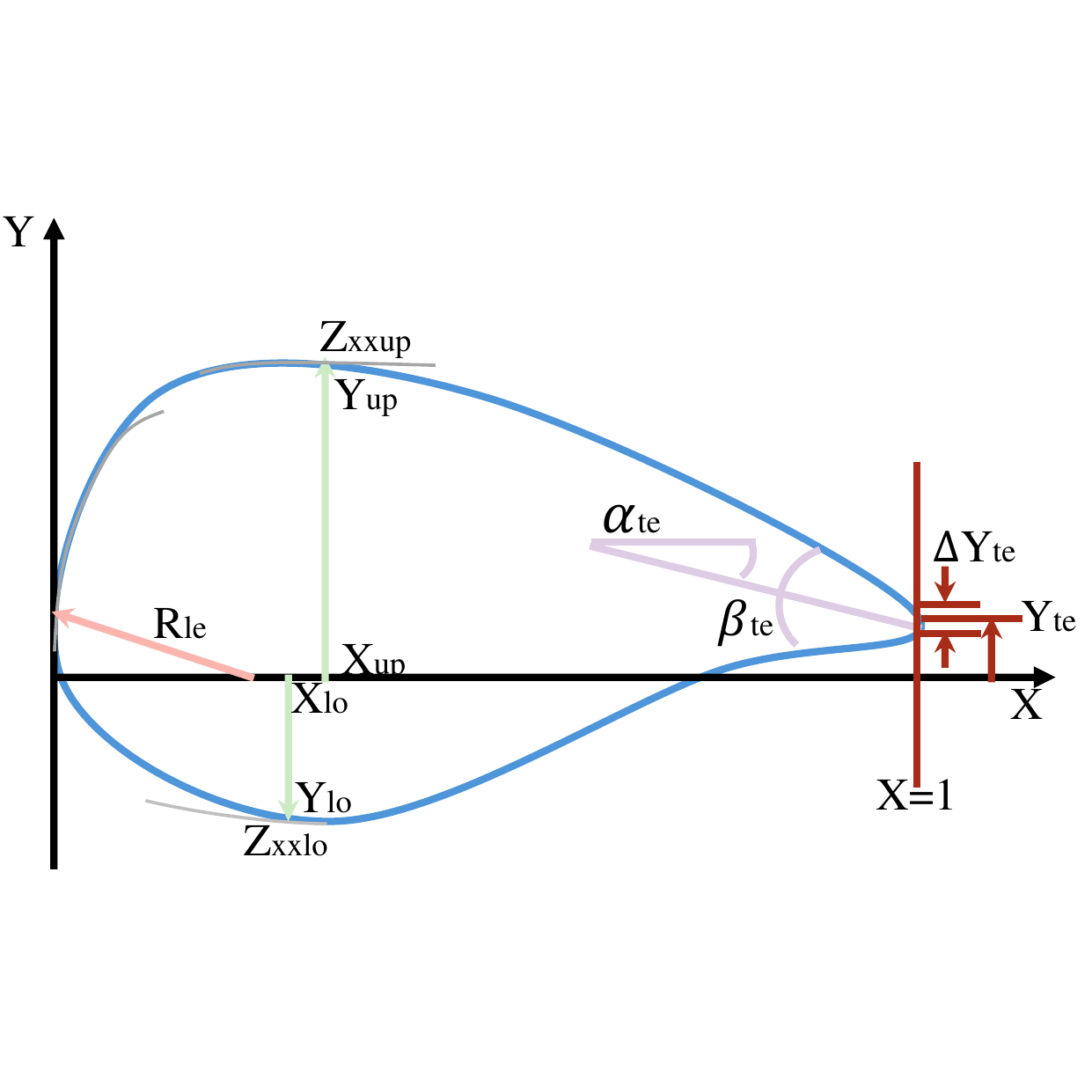}
    \caption{Demonstration of PARSEC parameters.}
    \label{fig:geoparams_PARSEC}
  \end{subtable}
  \caption{Eleven PARSEC parameterization in airfoil design.}
  \label{tab:geoparams}
\end{table}

\begin{algorithm}[t]
    \caption{Model Training.}
    \label{alg:pretrain}
    \footnotesize
    \textbf{Input}: data resolution $d$, data $u_1$, design condition variables $c$ (optional).\\
    \textbf{Parameter}: Gaussian process $\mathcal{GP}(0,K)$ for sampling $u_0$.\\
    \textbf{Output}: velocity operator $v_{\theta}$. 
    \begin{algorithmic}[1] 
        \WHILE{training...}
        \STATE sample $t\sim[0,1]$, $u_0\sim\mathcal{GP}(0,K)$ at resolution $d$ and get $\{u_{0,i}\}$.
        \STATE Compute $v_t$ at resolution $d$ and get $\{v_{t,i}\}$.
        \STATE Compute $u_t$ at resolution $d$ and get $\{u_{t,i}\}$.
        \STATE Compute $v_{\theta}(u_t,c,t)$ at resolution $d$ and get $\{v_{\theta,i}\}$.
        \STATE Minimize $\|\{v_{\theta,i}\}-\{v_{t,i}\}\|^2$.
        \STATE Compute gradient and update $\theta$.
        \ENDWHILE
        \STATE \textbf{return} $v_{\theta}$.
    \end{algorithmic}
\end{algorithm}

\begin{algorithm}[t]
    \caption{Model Inference.}
    \label{alg:generation}
    \footnotesize
    \textbf{Input}: sampling resolution $d$, sampling time steps $T$ and steps length $\mathrm{d}t$, latent function $u_0$ (optional), design condition variables $c$ (optional).\\
    \textbf{Parameter}: Gaussian process $\mathcal{GP}(0,K)$ for sampling $u_0$.\\
    \textbf{Output}: airfoil $\{y_i\}$ at resolution $d$. 
    \begin{algorithmic}[1] 
        \STATE Let $t=0$, $u_0\sim\mathcal{GP}(0,K)$ at resolution $d$ and get $\{u_{0,i}\}$.
        \WHILE{$t\leq 1$}
        \STATE Compute $v_{\theta}(u_t,c,t)$ at resolution $d$ and get $\{v_{\theta,i}\}$.
        \STATE Compute $\{u_{t+\mathrm{d}t,i}\}=\{u_{t,i}\}+\{v_{\theta,i}\mathrm{d}t\}$.
        \STATE $t=t+\mathrm{d}t$.
        \ENDWHILE
        \STATE \textbf{return} $\{y_i\}=\{u_{1,i}\}$
    \end{algorithmic}
\end{algorithm}

\begin{algorithm}[t]
    \caption{Model Finetuning (Airfoil Editing).}
    \label{alg:editing}
    \footnotesize
    \textbf{Input}: pretrained neural operator $v_{\theta}$, original airfoil function $u_1$ (optional) or latent function $u_0$ (optional), editing requirement $\delta$, editing resolution $d$, sampling time steps $T$ and steps length $\mathrm{d}t$.\\
    \textbf{Parameter}: Gaussian process $\mathcal{GP}(0,K)$ for sampling $u_0$, noise level $\sigma$.\\
    \textbf{Output}: new airfoil $\{y_i\}$ at resolution $d$. 
    \begin{algorithmic}[1] 
        \STATE Inversely sample original airfoil function $\{u_{1,i}\}$ through $v_{\theta}$ and get its latent function $\{u_{0,i}\}$.
        \STATE Set $\{a_{\theta,i}\} = \{u_{0,i}\}$.
        \WHILE{finetuning...}
        \STATE Sample $\{u_{1,i}^{'}\}$ through $v_{\theta}$ from $\{a_{\theta,i}\}$.
        \STATE Compute $\mathcal{L}_{\mathrm{MAP}}$.
        \STATE Compute gradient and update $\theta$.
        \ENDWHILE
        \STATE Sample $\{u_{1,i}^{'}\}$ through $v_{\theta}$ from $\{a_{\theta,i}\}$.
        \STATE \textbf{return} $\{y_i\}=\{u_{1,i}^{'}\}$
    \end{algorithmic}
\end{algorithm}

\begin{table}[t]
    \centering
    \caption{Training hyperparameters and model parameters in conditional airfoil generation tasks}
    \footnotesize
    \begin{tabular}{lc}
        \toprule
        \textbf{Training hyperparameters} & \textbf{Value} \\
        \midrule
        Max learning rate & $5\times10^{-6}$ \\
        Batch size  & $1024$ \\
        Batch size (AF-200K) & $2048$ \\
        Optimizer (pre-training and fine-tuning) & Adam \\
        Optimizer scheduler & Cosine Annealing \\
        Warmup iterations & $2000$ \\
        Max training iterations & $1000000$ \\
        Max training iterations (AF-200K) & $2000000$ \\
        ODE solver time steps & $10$ \\
        ODE solver type & Euler \\
        \midrule
        \textbf{Model parameters} & \textbf{Value} \\
        \midrule
        Fourier neural operator layers & $6$ \\
        Fourier neural operator modes & $64$ \\
        Fourier neural operator hidden channels & $256$ \\
        Gaussian process kernel & Matérn \\
        Kernel noise scale $n$ & $1$ \\
        Kernel order $\nu$ & $2.5$ \\
        Kernel characteristic length $l$ & $0.03$ \\
        \bottomrule
    \end{tabular}
    \label{tab:training_params}
\end{table}

We adopted the same design condition parameters as provided in AFBench~\cite{liu2024afbench}, as shown in Table~\ref{tab:geoparams}. The parameters listed below—including the leading-edge radius, the curvature at the maximum thickness of the upper and lower surfaces, and the trailing-edge angle of the airfoil—were obtained via curve fitting. Note that our implementation of the trailing-edge angle calculation slightly differs from that of AFBench, as we found the original approximation method not invariant to increased sampling precision. Specifically, AFBench employs B-spline curves for interpolation at the trailing edge and calculates its gradient at the endpoint, which becomes unstable as resolution increases. In contrast, our method assumes the last $2\%$ length of the airfoil curve near the trailing edge is nearly straight, and uses linear regression to determine the angle.

\section{Algorithm Implementation Details}\label{sec:appendix_algorithm}

The algorithm for training the airfoil generative model is presented in Algorithm~\ref{alg:pretrain}, and the inference procedure using a trained model is described in Algorithm~\ref{alg:generation}. In our implementation, we employed Optimal Flow Matching and utilized a Gaussian process with specific kernel functions, such as the RBF kernel or the Matérn kernel, in the latent space of the generative model.

\section{Training Details}
\label{app:hyper}

In most of the experiments, we followed the hyperparameters listed in the Table~\ref{tab:training_params} below to train the models, including both the pre-training and fine-tuning stages. On the AF-200K dataset, we increased the maximum number of iterations to $2000000$ and the batch size to $2048$. For the number of time steps used to solve the ODE, we use $10$ steps for relative balance between fast sampling speed and good performance, except for the fine-tuning stage, where we used 10 steps for training efficiency.

\begin{table}[t]
\centering
\caption{Time cost comparison of different generative models.}
\label{tab:cost}
\resizebox{\textwidth}{!}{
\begin{tabular}{lcccc}
\hline
\textbf{Model} & \textbf{\begin{tabular}[c]{@{}c@{}}Wall-clock for \\ 1 000 epochs \end{tabular}} & \textbf{\begin{tabular}[c]{@{}c@{}} NFEs at \\ test time\end{tabular}} & \textbf{\begin{tabular}[c]{@{}c@{}}Mean inference time \end{tabular}} & \textbf{\begin{tabular}[c]{@{}c@{}}GPU memory \\ at test\end{tabular}} \\ \hline
PK-DIT (score matching) & $\approx$ 10 h & 50 (DDIM) & 220 ms & $\approx$ 200 MB \\
FuncGenFoil (flow matching) & $<$ 6 h & 10 & 50 ms & $\approx$ 200 MB \\ \hline
\end{tabular}
}
\end{table}

To provide a comprehensive analysis of the model's computational requirements, we evaluate its performance relative to the point-based PK-DIT model. Table~\ref{tab:cost} details this comparison, presenting the wall-clock training time, per-sample inference latency, and peak GPU memory usage. All benchmarks were conducted on a single desktop machine equipped with an NVIDIA RTX 4090 and an Intel i9-13900K.

\begin{table}[t]
\centering
\caption{Effective ranges of the 11 geometric parameters for supercritical dataset.}
\label{tab:effective_range}
\begin{tabular}{clrr}
\hline
\textbf{Index} & \textbf{Parameter} & \textbf{Minimum} & \textbf{Maximum} \\ \hline
1  & leading edge radius      & 0.0073   & 0.0140   \\
2  & upper crest position x   & 0.3960   & 0.5200   \\ 
3  & upper crest position y   & 0.0592   & 0.0784   \\ 
4  & upper crest curvature    & -0.4580  & -0.2100  \\ 
5  & lower crest position x   & 0.3180   & 0.4100   \\
6  & lower crest position y   & -0.0589  & -0.0414  \\
7  & lower crest curvature    & 0.3730   & 0.8050   \\
8  & trailing edge position   & -0.0001  & 0.0001   \\
9  & trailing thickness       & 0.0020   & 0.0075   \\
10 & trailing edge angle up   & -0.5514  & -0.2254  \\
11 & trailing edge angle down & -0.4477  & -0.1397  \\
\hline
\end{tabular}
\end{table}

\section{More Experiments}

\subsection{Comparison with a Classical Optimization Method}\label{app:tradition_method}

To establish a performance benchmark, we compare our method against a classical, optimization-based approach for airfoil reconstruction. This benchmark is designed to estimate the near-optimal accuracy achievable with a standard parametric representation.

The classical method represents the airfoil using a NURBS curve with $24$ control points. To create a best-case scenario for this method, the optimization for each of the $3825$ target airfoils in the test dataset was initialized with control points configured to match the \textit{average} airfoil shape. This starting point is already very close to the final target, ensuring stable and rapid convergence. For each target, the optimization was performed using the Adam optimizer for 100 steps with a learning rate of $3\times10^{-4}$, continuing until the Mean Squared Error (MSE) plateaued. The total computation cost for fitting the entire test set was approximately 20 GPU-hours on an NVIDIA A800 GPU.

The results of this comparison are summarized in Table~\ref{tab:tradition_method}. As expected, the NURBS fitting baseline performs exceptionally well, closely approaching what might be considered a practical upper bound on accuracy for a $24$-parameter representation. FuncGenFoil's accuracy is competitive with this strong baseline. Analyzing the geometric errors reveals that the NURBS fitting method excels at the trailing edge, likely because the control points can effectively constrain this region. However, FuncGenFoil demonstrates superior performance at the leading edge. The classical method struggles with the high curvature of the leading edge, where the $24$ control points lack the flexibility to capture the geometry accurately.

It is crucial to highlight a fundamental difference in the problem setup that heavily favors the classical method. Its optimization begins from a well-conditioned starting point, whereas generative models like FuncGenFoil or PK-DIT must learn to construct airfoils from a state of maximum entropy (i.e., pure Gaussian noise)—a significantly more challenging task. We note that the classical optimization process is delicate; initializing the control points more randomly often caused the optimization to become unstable and diverge.

Therefore, this experiment should not be interpreted as a direct comparison. Instead, it provides a reference for the best-case reconstruction performance of a model-free parametric method under ideal optimization conditions.

\begin{table*}[t]
    \centering
    \caption{Performance Comparison: Classical Optimization vs. FuncGenFoil. The classical "NURBS fitting" method is initialized close to the target, representing a near-optimal benchmark.}
    \label{tab:tradition_method}
    \resizebox{\textwidth}{!}{
    \begin{tabular}{ccccccccccccccc}
        \toprule
        \multirow{2}{*}{\textbf{Dataset}}
        & \multirow{2}{*}{\textbf{Algo.}}
        & \multicolumn{13}{c}{\textbf{Label Error ($\times 10^{-3}$) $\downarrow$}} 
        \\ \cline{3-15}
        & & $\sigma_1$ & $\sigma_2$ & $\sigma_3$ & $\sigma_4$ & $\sigma_5$ & $\sigma_6$ & $\sigma_7$ & $\sigma_8$ & $\sigma_9$ & $\sigma_{10}$ & $\sigma_{11}$ & $\bar{\sigma}_a$ & $\bar{\sigma}_g$ \\ \hline
        \multirow{2}{*}{Super} 
        & NURBS fitting
        & 10.8 & 17.4 & 4.5 & 64.5 & 9.79 & 3.56 & 113.2 & 1.53 & 1.72 & 0.065 & 0.106 & 20.6 & 3.97 \\
        & FuncGenFoil
       & 0.71 & 8.23 & 0.13 & 201.3 & 4.72 & 0.12 & 174.2 & 0.09 & 0.14 & 34.2 & 36.7 & 41.9 & 3.08  \\
        \bottomrule
    \end{tabular}
    }
\end{table*}

\subsection{More Ablations Experiments}\label{app:more_ablations}

We conduct a comprehensive ablation study on the choice of Gaussian process kernel types, model modes of Fourier Neural Operators, and characteristic lengths, as shown in Table~\ref{tab:ablation_matern} for the Matérn kernel, Table~\ref{tab:ablation_RBF} for the radial basis function (RBF) kernel, and Table~\ref{tab:ablation_white} for the white noise kernel, which is commonly used in diffusion or flow models in finite-dimensional spaces. In general, appropriate selections of these settings can significantly enhance model performance. Specifically, the Gaussian prior should exclude function spaces where the data is unlikely to reside, making the prior space as restricted as possible to facilitate effective model training. For example, a dataset with high smoothness is unlikely to belong to the function space generated by a white noise prior but is more plausibly sampled from a Gaussian prior with an RBF kernel.

In Table~\ref{tab:ablation_RBF}, we observe that the RBF kernel exhibits a similar tendency to the Matérn kernel, with both achieving optimal performance at the characteristic length $0.03$. As for the white noise kernel in Table~\ref{tab:ablation_white}, since its characteristic length is effectively $0$, we studied the influence of the noise scale $n$ over the range $[1.0, 2.0, 4.0]$. We found that excessively large diffusion white noise negatively impacts model performance, likely because such noise scales project the function into a larger white noise space, making it more difficult for the neural operator to learn the intrinsic dynamics.

\begin{table*}[t]
  \centering
  \renewcommand{\arraystretch}{1.2}
  \setlength{\tabcolsep}{4pt}
  \caption{FuncGenFoil model performance with different FNO mode and Matérn kernel for conditional airfoil generation.}
  \label{tab:ablation_matern}
  \resizebox{\textwidth}{!}{%
    \begin{tabular}{c c c *{13}{c} c c}
      \toprule
      \multirow{2}{*}{\textbf{Modes}} 
        & \multirow{2}{*}{\textbf{$\nu$}} 
        & \multirow{2}{*}{\textbf{$l$}} 
        & \multicolumn{13}{c}{\textbf{Label Error \boldmath$\downarrow (10^{-3})$}} 
        & \multirow{2}{*}{\boldmath$\mathcal{D}\uparrow$} 
        & \multirow{2}{*}{\boldmath$\mathcal{M}\downarrow (10^{-2})$} \\
      \cline{4-16}
      & & 
        & $\sigma_1$ & $\sigma_2$ & $\sigma_3$ & $\sigma_4$ & $\sigma_5$ 
        & $\sigma_6$ & $\sigma_7$ & $\sigma_8$ & $\sigma_9$ & $\sigma_{10}$ 
        & $\sigma_{11}$ & $\bar{\sigma}_a$ & $\bar{\sigma}_g$ 
        & & \\
      \midrule
      \multirow{6}{*}{$32$} 
        & \multirow{3}{*}{$2.5$} 
        & $0.01$    
          &  0.62  &  6.06  &  $\mathbf{0.09}$  & 200.4  & 3.58   
          &  0.09 & 185.3  &  $\mathbf{0.07}$  & 0.12   & 26.7   
          & 29.1  & 41.1    & 2.51   & -105.6  &  1.02 \\
        & 
        & $0.02$ 
          & 0.52  & 5.10   & 0.10   & 97.4  &  3.26  
          & 0.11  & 107.0  &  0.08  & 0.12   &  26.2  
          & $\mathbf{27.6}$   & 24.3    & 2.22   & -105.7  &  $\mathbf{0.99}$ \\
        & 
        & $0.03$     
          &  0.54 &  $\mathbf{4.78}$  &  0.11  & $\mathbf{82.3}$  & 3.00   
          &  0.11 &  93.2 &  0.09  & 0.12  &  $\mathbf{26.0}$  
          & 27.7   &  $\mathbf{21.6}$   & $\mathbf{2.18}$   & -105.6  &  $\mathbf{0.99}$ \\
        \cline{2-18}
        & \multirow{3}{*}{$3.5$} 
        & $0.01$   
          &  0.63 &  5.64  &  $\mathbf{0.09}$  & 158.3  & 3.46   
          & $\mathbf{0.08}$  &  156.9 &  $\mathbf{0.07}$  &  0.12  &  27.6  
          &  29.2  &  34.7   & 2.41   & -105.8  &  1.01 \\
        & 
        & $0.02$ 
          &  $\mathbf{0.48}$ &  5.08  &  0.10  &  89.4 &  3.12  
          &  0.10 &  97.5 &  0.08  &  0.12  &  26.8  
          &  28.2  & 22.8    & $\mathbf{2.18}$   & -105.9  &  $\mathbf{0.99}$  \\
        & 
        & $0.03$      
          &  0.53 &  5.17  &  0.11  &  83.5 &  $\mathbf{2.90}$  
          &  0.11 &  $\mathbf{89.5}$ &  0.10  &  0.14  &  29.4  
          &  30.7  &  22.0   & 2.28   & -105.2  &  $\mathbf{0.99}$  \\
      \midrule
      \multirow{12}{*}{$64$} 
        & \multirow{3}{*}{$1.5$} 
        & $0.01$    
          &  0.82 &  8.57  &  0.10  &  370.2 & 5.21   
          &  0.09 &  342.5 &  0.08  &  0.12  &  36.3  
          &  37.5  &  72.9   & 3.30   & -104.0  & 1.09  \\
        & 
        & $0.02$ 
          &  0.65 &  8.26  &  0.13  &  222.9 & 4.97   
          & 0.11  & 219.6  &  0.08  &  0.13  &  37.4  
          &  35.9  &  48.2   & 3.12   & -103.6  &  1.04 \\
        & 
        & $0.03$     
          &  0.71 & 8.18   &  0.14  &  183.7 &  4.82  
          &  0.12 &  207.2 & 0.09   &  0.14  &  34.6  
          &  36.6  &  43.3   & 3.13   & -104.0  & 1.02  \\
        \cline{2-18}
        & \multirow{3}{*}{$2.5$} 
        & $0.01$    
          &  0.80 &  8.77  & 0.10   & 291.2  & 5.23   
          & 0.10  & 277.6 &  0.08  & 0.12   &  35.8  
          &  37.7  &  59.8   & 3.20   & -103.7  & 1.06  \\
        & 
        & $0.02$ 
          &  0.68 & 8.66  &  0.13  &  192.4 &  4.78  
          &  0.11 &  182.8 &  0.09  &  0.12  &  36.2  
          & 37.8   & 42.2    & 3.06   & -103.5  &  1.02 \\
        & 
        & $0.03$     
          & 0.71 & 8.23 & 0.13 & 201.3 & 4.72 & 0.12 & 174.2 & 0.09 & 0.14 & 34.2 & 36.7 & 41.9 & 3.08 & -103.9 & 1.01 \\
        \cline{2-18}
        & \multirow{6}{*}{$3.5$} 
        & $0.01$   
          &  0.88 & 8.80   &  0.11  &  269.3 &  5.26  
          &  0.09 &  248.9 &  0.08  &  0.12  &  36.5  
          &  38.0  &  55.3   & 3.20   & -103.4  &  1.05 \\
        & 
        & $0.02$ 
          &  0.72 &  8.52  & 0.11   & 177.1  &  4.84  
          &  0.11 & 174.6  &  0.08  &  0.13  &  33.2  
          &  37.4  &  39.7   & 2.98   & -103.6  & 1.02  \\
        & 
        & $0.03$      
          &  0.66 & 7.80   & 0.12   &  161.1 &  4.54  
          &  0.11 &  164.2 &  0.09  &  0.15  &  35.1  
          &  38.3  &  37.5   & 2.96   & -103.4  &  1.01 \\
        &
        & $0.04$   
          & 0.82 & 8.19  & 0.13  & 170.0  & 4.45  
          & 0.13 & 159.7 & 0.09  & 0.16  & 36.6
          & 34.8  & 37.7   & 3.11   & -103.6 & 1.01 \\
        & 
        & $0.06$ 
          & 0.87 & 8.82  & 0.16  & 183.0 & 5.01
          & 0.13 & 183.9 & 0.10  & 0.17  & 41.8  
          & 41.2  & 42.3   & 3.44   & -102.8 & 1.02 \\
        & 
        & $0.12$      
          & 1.39 & 10.7  & 0.20  & 261.0 & 6.23  
          & 0.15 & 230.3 & 0.12  & 0.19  & 53.4 
          & 44.9  & 55.3   & 4.27   & $\mathbf{-99.9}$ & 1.05 \\
      \midrule
      \multirow{6}{*}{$128$} 
        & \multirow{3}{*}{$2.5$} 
        & $0.01$    
          & 1.09 & 10.4  & 0.11  & 708.5 & 7.13  
          & 0.10 & 613.5 & 0.09  & $\mathbf{0.11}$  & 40.7 
          & 41.1  & 129.3   & 4.16   & -102.8 & 1.27 \\
        & 
        & $0.02$ 
          & 1.15 & 10.0  & 0.11  & 577.5 & 6.83  
          & 0.11 & 570.2 & 0.09  & 0.14  & 37.8  
          & 42.3  & 113.3   & 4.14   & -103.4 & 1.20 \\
        & 
        & $0.03$     
          & 1.05 & 10.1  & 0.12  & 598.0 & 6.48 
          & 0.14 & 487.7 & 0.09  & 0.15  & 36.9  
          & 40.1  & 107.3   & 4.22   & -103.0 & 1.23 \\
        \cline{2-18}
        & \multirow{3}{*}{$3.5$} 
        & $0.01$   
          & 1.04 & 10.3 & 0.10  & 628.9 & 7.05  
          & 0.09 & 576.8 & 0.08  & 0.12  & 39.3  
          & 42.0  & 118.7   & 4.05   & -102.9 & 1.23 \\
        & 
        & $0.02$ 
          & 0.93 & 9.94  & 0.12  & 560.6 & 6.81  
          & 0.10 & 484.6 & 0.09  & 0.15  & 41.5  
          & 40.8  & 104.1 & 4.05  & -103.2 & 1.19 \\
        & 
        & $0.03$      
          & 1.11 & 9.93  & 0.11 & 548.2 & 6.49  
          & 0.12 & 520.3 & 0.09 & 0.16  & 38.6  
          & 39.0  &  105.8  & 4.16  & -103.1 & 1.23 \\
      \bottomrule
    \end{tabular}%
  }
\end{table*}

\begin{table*}[t]
    \centering
    \caption{FuncGenFoil model performance with different RBF kernel for conditional airfoil generation.}
    \label{tab:ablation_RBF}
    \resizebox{\textwidth}{!}{
    \begin{tabular}{ccccccccccccccccc}
        \toprule
        \multirow{2}{*}{\textbf{Modes}}
        & \multirow{2}{*}{\textbf{$l$}}
        & \multicolumn{13}{c}{\textbf{Label Error \boldmath$\downarrow (10^{-3})$}} 
        & \multirow{2}{*}{\boldmath$\mathcal{D} \uparrow$} 
        & \multirow{2}{*}{\boldmath$\mathcal{M} \downarrow (10^{-2})$} 
        \\ \cline{3-15}
 & & $\sigma_1$ & $\sigma_2$ & $\sigma_3$ & $\sigma_4$ & $\sigma_5$ & $\sigma_6$ & $\sigma_7$ & $\sigma_8$ & $\sigma_9$ & $\sigma_{10}$ & $\sigma_{11}$ & $\bar{\sigma}_a$ & $\bar{\sigma}_g$ &  \\ \hline
        \multirow{7}{*}{$64$} 
        & $0.001$ & 1.26 & $\mathbf{7.60}$  &  $\mathbf{0.09}$  & 1224 & 5.34 &$\mathbf{0.07}$ & 1046 & $\mathbf{0.06}$ & $\mathbf{0.08}$ &  24.9 &$\mathbf{28.6}$ & 213 &3.68 &  -103.7 & 1.42\\
        & $0.01$ & 0.77 & 8.67 & 0.12 & 223.6 & 5.05 & 0.09 & 174.7 & 0.08 & 0.12& 35.2 & 37.1 & 47.0 & 3.06 & -103.6 & 1.03 \\
        & $0.02$ &  $\mathbf{0.62}$  & 8.33 & 0.12 & 162.2 & 4.59 & 0.11 & 165.9  & 0.09 & 0.17 & 38.0 & 42.5 & 38.4 & 3.04 & -103.2 &  $\mathbf{1.01}$ \\
        & $0.03$ & 0.66 & 7.80 & 0.12 &  $\mathbf{147.9}$ &  $\mathbf{4.43}$ & 0.12 &  $\mathbf{154.2}$ & 0.09 & 0.15 & $\mathbf{33.2}$ & 37.9 &  $\mathbf{35.1}$& $\mathbf{2.93}$ & -103.6 & 1.00 \\
        & $0.04$ & 0.84 & 8.79 & 0.13 & 197.1 & 4.59 & 0.12 & 168.4 & 0.09 & 0.17 & 35.8 & 37.9 & 41.3 & 3.20 & -103.2 & 1.02 \\
        & $0.06$ & 1.15 & 9.97 & 0.15 & 220.3 & 6.21 & 0.14 & 215.5 & 0.10 & 0.16 & 38.3 & 44.2 & 48.7 & 3.73 & -99.9 & 1.05  \\
        & $0.12$ & 1.12 & 10.9 & 0.17 & 230.5 & 7.71 & 0.17 & 265.6 & 0.11 & 0.18 & 40.0 & 42.3 & 54.4 & 4.14 & $\mathbf{-95.1}$ & 1.08  \\
        \bottomrule
    \end{tabular}
    }
\end{table*}

\begin{table*}[t]
    \centering
    \caption{FuncGenFoil model performance with white noise kernel of different noise scale $n$ for conditional airfoil generation.}
    \label{tab:ablation_white}
    \resizebox{\textwidth}{!}{
    \begin{tabular}{ccccccccccccccccc}
        \toprule
        \multirow{2}{*}{\textbf{Modes}}
        & \multirow{2}{*}{\textbf{$n$}}
        & \multicolumn{13}{c}{\textbf{Label Error \boldmath$\downarrow (10^{-3})$}} 
        & \multirow{2}{*}{\boldmath$\mathcal{D} \uparrow$} 
        & \multirow{2}{*}{\boldmath$\mathcal{M} \downarrow (10^{-2})$} 
        \\ \cline{3-15}
 & & $\sigma_1$ & $\sigma_2$ & $\sigma_3$ & $\sigma_4$ & $\sigma_5$ & $\sigma_6$ & $\sigma_7$ & $\sigma_8$ & $\sigma_9$ & $\sigma_{10}$ & $\sigma_{11}$ & $\bar{\sigma}_a$ & $\bar{\sigma}_g$ &  \\ \hline
        \multirow{3}{*}{$64$} 
        & 1
        & 1.15 & 7.18 & 0.09 & 913.8 & 5.16 & 0.07 & 964.5 & 0.06 & 0.08 & 26.2 & 26.1 & 176.8 & 3.44 & -104.7 & 1.39 \\
        & 2
       & 1.46 & 7.98 & 0.10 & 1274 & 5.94 & 0.08 & 1268 & 0.08 & 0.10 & 30.6 & 32.4 & 238.3 & 4.17 & -102.7 & 1.64 \\
        & 4 
        & 2.27 & 9.14 & 0.12 & 2015 & 7.24 & 0.11 & 2059 & 0.09 & 0.14 & 41.5 & 41.4 & 379.7 & 5.61 & -99.9 & 2.09 \\
        \bottomrule
    \end{tabular}
    }
\end{table*}

\subsection{Generated Airfoil Examples}

\begin{figure}[t]
\small
\centering
\includegraphics[width=\linewidth]{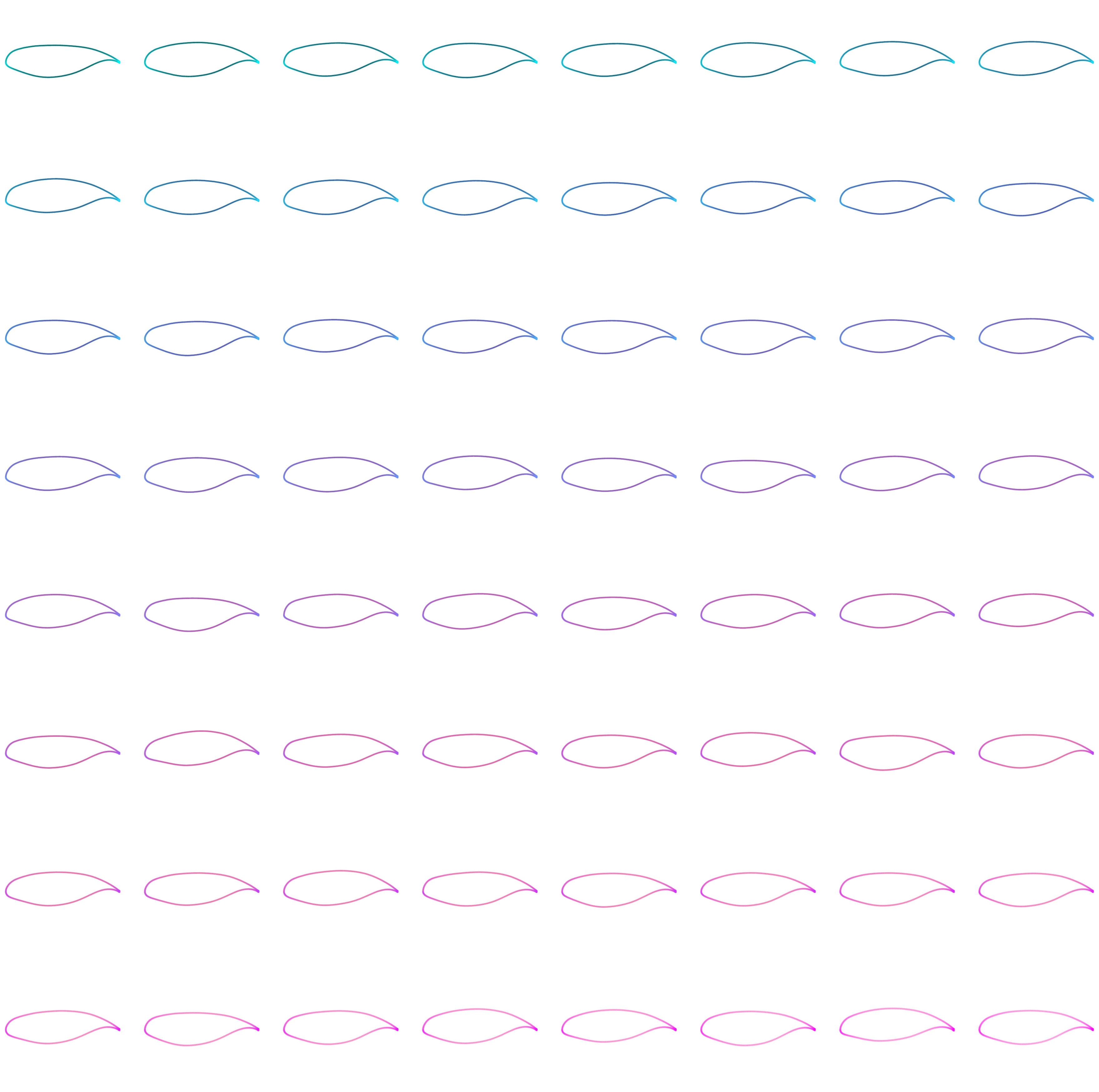}
\caption{A diverse set of airfoil geometries generated by FuncGenFoil models trained on supercritical airfoil data.}
\label{fig:demo}
\end{figure}

Figure~\ref{fig:demo} showcases a variety of airfoil geometries generated by FuncGenFoil models. These models were trained on datasets of supercritical airfoils, demonstrating their capability to produce diverse and plausible shapes.

\section{Societal Impacts}\label{app:societal_impacts}

This work will accelerate research in AI for Science and Engineering and generally has positive social impacts. On the other hand, AI models for engineering design and optimization could accelerate technological diffusion within society, potentially raising issues related to intellectual property or unauthorized technology transfer to entities or individuals intending to design specific engineering products or construct items harmful to society.

\end{document}